\documentclass[lettersize,journal]{IEEEtran}
\usepackage{amsmath,amsfonts}
\usepackage{algorithmic}
\usepackage{algorithm}
\usepackage{array}
\usepackage[caption=false,font=normalsize,labelfont=sf,textfont=sf]{subfig}
\usepackage{textcomp}
\usepackage{stfloats}
\usepackage{url}
\usepackage{verbatim}
\usepackage{graphicx}
\usepackage{cite}
\usepackage{multirow}
\usepackage{bbm}
\usepackage{nicefrac}
\usepackage{xcolor}

\usepackage{booktabs}

\begin{document}

\title{Towards Better Performance in Incomplete LDL: Addressing Data Imbalance}

\author{Zhiqiang~Kou, Haoyuan~Xuan, ~Jing~Wang,~Yuheng~Jia,~\IEEEmembership{Member,~IEEE,}\\and~Xin Geng,~\IEEEmembership{Senior~Member,~IEEE}
	\thanks{Manuscript received April 19, 2021; revised August 16, 2021. This work was supported  in part by  the  National Natural Science Foundation of China under Grant 62125602, Grant 62076063, Grant 62106044 and Grant 62306073,  in part by the Natural Science Foundation of Jiangsu Province under Grant BK20210221 and Grant BK20230832, in part by  the China Postdoctoral Science Foundation under Grant 2022M720028, and in part by ZhiShan Youth Scholar Program from Southeast University under Grant 2242022R40015. (\textit{Zhiqiang Kou and Haoyuan Xuan contributed equally to this work}) (\textit{Corresponding authors:Yuheng Jia and  Xin Geng}).}
	\thanks{
		The authors are with the School of Computer Science and Engineering, Southeast University, Nanjing 211189, China, and also with the Key Laboratory of New Generation Artificial Intelligence Technology and Its Interdisciplinary Applications (Southeast University), Ministry of Education, China. Yuheng Jia is also with the School of Computing \& Information Sciences, Caritas Institute of Higher Education, Hong Kong (e-mail: zhiqiang\_kou@seu.edu.cn; 230238557@seu.edu.cn; dongdongwu1230@gmail.com;  wangjing91@seu.edu.cn; yhjia@seu.edu.cn; 	zhangml@seu.edu.cn;  xgeng@seu.edu.cn).}
}
\markboth{Journal of \LaTeX\ Class Files,~Vol.~14, No.~8, August~2021}%
{Shell \MakeLowercase{\textit{et al.}}: A Sample Article Using IEEEtran.cls for IEEE Journals}


\maketitle

\begin{abstract}
	Label Distribution Learning (LDL) is a novel machine learning paradigm that addresses the problem of label ambiguity and has found widespread applications. Obtaining complete label distributions in real-world scenarios is challenging, which has led to the emergence of Incomplete Label Distribution Learning (InLDL).  However, the existing InLDL methods  overlook a crucial aspect of LDL data: the inherent imbalance in label distributions. To address this limitation, we propose \textbf{Incomplete and Imbalance Label Distribution Learning (I\(^2\)LDL)}, a framework that simultaneously handles incomplete labels and imbalanced label distributions. Our method decomposes the label distribution matrix into a low-rank component for frequent labels and a sparse component for rare labels, effectively capturing the structure of both head and tail labels. We optimize the model using the Alternating Direction Method of Multipliers (ADMM) and derive generalization error bounds via Rademacher complexity, providing strong theoretical guarantees. Extensive experiments on 15 real-world datasets demonstrate the effectiveness and robustness of our proposed framework compared to existing InLDL methods.
\end{abstract}

\section{Introduction}

 \IEEEPARstart{I}{n} most supervised machine learning tasks, the goal is to establish a mapping from the feature space to the label space to predict which candidate labels are associated with a given instance. However, in real-world scenarios, the distinction between relevant and irrelevant labels is often ambiguous. Traditional learning frameworks, such as Single-label Learning (SLL) and Multi-label Learning (MLL)\cite{zhang2013review, tsoumakas2007multi}, treat all candidate labels as equally important, which is not always practical for many applications. This challenge is commonly referred to as learning with ambiguity\cite{gao2017deep}.  To address this limitation, Geng\cite{geng2016label} introduced the Label Distribution Learning (LDL) framework\cite{jia2018label,jia2019label,jia2023adaptive,wang2021label_2, WangFL}, which extends SLL and MLL by associating each instance with a numerical vector known as the label distribution. Each component of this vector reflects the degree of relevance of a particular label to the instance, offering a more nuanced representation than traditional approaches\cite{kou}. 
 LDL has been successfully applied to a variety of real-world tasks, including facial age estimation\cite{wen2020adaptive}, soft facial landmark detection\cite{su2019soft}, head pose estimation\cite{geng2020head}, facial expression recognition\cite{chen2020label}, and historical context-based style classification\cite{yang2018historical}.

 Obtaining complete label degrees is often laborious and challenging in real-world scenarios \cite{KOUCSVT}. This difficulty has driven the development of Incomplete LDL (InLDL)\cite{xu2017incomplete, zhang2022safe, teng2021incomplete}, as researchers seek to overcome such obstacles. Existing research has shown that LDL data frequently exhibits significant imbalance\cite{zhao2023imbalanced}, where certain labels are overrepresented while others, particularly tail labels, are relatively rare, resulting in poor predictive performance for these underrepresented labels. For example, in models predicting movie rating distributions\cite{geng2015pre}, biases in data collection may cause certain movie genres to receive disproportionately higher description degrees for specific ratings, leading to imbalanced label distributions.  When label distributions are randomly missing, as in the InLDL setting, the already sparse data for tail labels becomes even scarcer, leading to a further decline in the model's predictive performance for these labels\cite{xu2017incomplete, wei2021towards}. We further demonstrate this effect in our theoretical analysis. Unfortunately, most current InLDL methods operate under the assumption of balanced data and fail to address this inherent label imbalance. This gap highlights the pressing need for a new method capable of effectively handling both incomplete and imbalanced label distributions.

 To overcome the limitations of current InLDL methods, which often fail to account for the inherent label imbalance present in real-world data, we propose a novel framework called \textbf{Incomplete and Imbalance Label Distribution Learning (I\(^2\)LDL)}. Our approach simultaneously addresses both incomplete labels and imbalanced label distributions, building on the observation that real-world label distributions are often long-tailed, with certain labels significantly underrepresented. This imbalance worsens predictive performance on tail labels, especially in the presence of incomplete data. The core of our method involves decomposing the label distribution matrix into two components: a low-rank matrix that captures the structure of the frequent, or "head," labels, and a sparse matrix that represents the rare, or "tail," labels. By leveraging low-rank approximations, we effectively model the more common labels, while sparsity constraints ensure sensitivity to the underrepresented tail labels. This decomposition enables \textit{I\(^2\)LDL} to provide a more robust and accurate representation of label distributions, even in challenging, real-world datasets. Finally, we solve the resulting optimization problem using the Alternating Direction Method of Multipliers (ADMM) \cite{boyd2011distributed}, ensuring efficient and scalable performance.

 To validate the robustness of our I\(^2\)LDL framework, we further establish a generalization error bound by leveraging Rademacher complexity, providing strong theoretical guarantees for handling both incomplete and imbalanced label distributions effectively. Additionally, extensive experiments on 15 real-world label distribution datasets with varying degrees of missingness demonstrate the effectiveness of the proposed I\(^2\)LDL framework.

 To sum up, our major contributions are summarized as follows:
 
 \begin{itemize}
 	\item  We propose a novel framework, \textbf{Incomplete and Imbalance Label Distribution Learning (I\(^2\)LDL)}, which simultaneously addresses the challenges of incomplete labels and imbalanced label distributions in label distribution learning tasks.

 	\item  We provide a comprehensive theoretical analysis by deriving generalization error bounds using Rademacher complexity. This analysis offers strong theoretical guarantees for the effectiveness of our method in handling both incomplete and imbalanced label distributions.
 	
 	\item  Extensive experiments conducted on 15 real-world label distribution datasets demonstrate the superior performance of our I\(^2\)LDL framework compared to existing Incomplete LDL methods.
 \end{itemize}
 
 \section{RELATED WORK}
 \subsection{Label Distribution Learning}

 LDL introduces label distributions as a novel learning paradigm to quantify the relevance of each label, drawing significant interest from researchers. This section provides a concise review of LDL research. LDL methods \cite{wangPLD, WangNN } are generally classified into three categories\cite{geng2016label}: problem transformation (PT), algorithm adaptation (AA), and specialized algorithm (SA). In PT, works like Geng \cite{geng2014multilabel} and Borchani et al. \cite{borchani2015survey} recast the LDL challenge as a single-label task using label probabilities as weights. AA methods modify traditional classifiers to meet LDL's unique needs, such as AA-kNN \cite{geng2016label}, which leverages neighbor distances to estimate label distributions. SA approaches often employ custom algorithms; for instance,
 LDL-DA \cite{10634763} address the issue of insufficient target data by learning shared representations and minimizing the distance between label prototypes, and LDL-SCL \cite{zheng2018label} improves prediction accuracy by utilizing local sample correlations. LDL-DPA \cite{10086552} designed to enhance the prediction of key labels by integrating a new metric. However, these LDL algorithms all assume that the label space is complete, whereas in reality, collecting a complete label distribution is undoubtedly challenging.

 \subsection{Incomplete Label Distribution Learning}
 
 In recent years, several methods have been proposed to address the problem of incomplete label distribution learning (InLDL), where complete label distributions are not always available for each instance. Xu et al. \cite{xu2017incomplete} introduced one of the earliest approaches, which handles missing labels by utilizing the neighborhood relationship among instances to infer missing label degrees. Teng et al. \cite{teng2021incomplete} further developed a matrix completion-based method that reconstructs missing label distributions by exploiting both feature and label correlations. Zhang et al. \cite{zhang2022safe} extended this concept by incorporating a safe learning framework, ensuring robustness against incomplete label distributions. However, these methods often overlook the label imbalance present in real-world data, focusing solely on the incompleteness aspect. Addressing both incompleteness and label imbalance remains a significant challenge, and existing InLDL methods fail to capture this duality, leading to suboptimal performance on underrepresented or tail labels.

\section{The PROPOSED METHOD}

\subsection*{3.1 Formalization}

In the framework of Incomplete Label Distribution Learning (IncomLDL), let \( \mathbf{X} = [\mathbf{x}_1; \mathbf{x}_2; \cdots; \mathbf{x}_n] \in \mathbb{R}^{n \times d} \) denote the feature space, where \( d \) is the dimension of the feature space, \( n \) is the number of instances, and \( \mathbf{x}_i \) represents the feature vector of the \(i\)-th instance. Let \( \mathbf{D} = [\mathbf{D}_1; \mathbf{D}_2; \cdots; \mathbf{D}_n] \in \mathbb{R}^{n \times m} \) be the label space, where \( \mathbf{D}_i = [d^{y_1}_{x_i}, d^{y_2}_{x_i}, \cdots, d^{y_m}_{x_i}] \) is the label distribution related to instance \( \mathbf{x}_i \). The term \( d^{y_j}_{x_i} \), called the description degree, represents the importance of label \( y_j \) for the instance \( \mathbf{x}_i \), which satisfies \( d^{y_j}_{x_i} \in [0,1] \) and \( \sum_{j=1}^{m} d^{y_j}_{x_i} = 1 \), where \( m \) is the total number of labels.

In the IncomLDL scenario, we assume that some elements in the label distribution matrix \( \mathbf{D} \) are missing at random. We denote the observed label distribution matrix by \( \tilde{\mathbf{D}} \), which is constructed by retaining certain observed elements from \( \mathbf{D} \) while setting the unobserved positions to 0. The incomplete label distribution matrix \( \tilde{\mathbf{D}} \) has the same dimensions as \( \mathbf{D} \), with observed entries identical to those in \( \mathbf{D} \) and unobserved entries set to zero. To indicate the observed entries, we define the binary matrix \( \Omega \in \mathbb{R}^{n \times m} \), where each element \( \Omega_{i,j} \) is defined as:

\[
\Omega_{i,j} = 
\begin{cases} 
	1, & \text{if } \tilde{D}_{i,j} \text{ is observed}, \\
	0, & \text{if } \tilde{D}_{i,j} \text{ is missing}.
\end{cases}
\]

Previous research \cite{zhao2023imbalanced} has demonstrated that LDL data inherently exhibits an imbalanced distribution, where some labels are overrepresented while others are underrepresented. In the current IncomLDL setting\cite{xu2017incomplete, teng2021incomplete, wang2021label_2}, where labels are randomly missing, this imbalance is further exacerbated, disproportionately impacting the minority labels.

To address this, we first provide a theoretical analysis showing how IncomLDL performance degrades as data imbalance worsens. We prove that the combination of imbalance and missing data significantly impacts performance, a problem that existing methods have not fully considered.

\section{Theoretical Analysis}

We first demonstrate how label imbalance, combined with random missing data, leads to performance degradation in Label Distribution Learning (LDL) models. The following analysis explains this degradation in terms of generalization error.

In LDL, for each label $y_i$, the model predicts a description degree $\hat{d}_i$ to approximate the true degree $d_i$. The generalization error for label $y_i$ can be decomposed as:

\[
\mathbb{E}[(\hat{d}_i - d_i)^2] = \text{Bias}^2(d_i) + \text{Variance}(d_i)
\]

In Incomplete Label Distribution Learning (IncomLDL), where labels are Missing At Random (MAR), imbalance is further exacerbated, disproportionately impacting minority labels. Let $\tilde{d}_i$ denote the observed description degree after accounting for missing data. The error becomes:

\[
\mathbb{E}[(\hat{d}_i - d_i)^2] = \mathbb{E}[(\hat{d}_i - \tilde{d}_i)^2] + \mathbb{E}[(\tilde{d}_i - d_i)^2]
\]

The second term, $\mathbb{E}[(\tilde{d}_i - d_i)^2]$, reflects the error due to missing labels. If $\beta_i$ denotes the missing rate for label $y_i$, the effective sample size is $N_i(1 - \beta_i)$, where $N_i$ is the original sample size. Thus, the generalization error bound becomes:

\[
\mathbb{E}[(\tilde{d}_i - d_i)^2] \leq O\left( \sqrt{\frac{d_i}{N_i (1 - \beta_i)}} \right)
\]

This shows that as $\beta_i$ increases, the effective sample size decreases, leading to higher error, especially for minority labels with smaller $N_i$. The overall generalization error across all labels is:

\[
\mathbb{E}[E_{\text{LDL}}] \leq \frac{1}{L} \sum_{i=1}^L \left( \mathbb{E}[(\hat{d}_i - \tilde{d}_i)^2] + O\left( \sqrt{\frac{d_i}{N_i (1 - \beta_i)}} \right) \right)
\]

In summary:
\begin{itemize}
	\item Label imbalance, combined with missing data, disproportionately increases the generalization error for minority labels.
	\item The overall error grows as missing data exacerbates the imbalance, degrading model performance in LDL.
\end{itemize}

\subsection{Proposed $I^2LDL$}

IncomLDL methods face challenges with both missing data and imbalanced label distributions. Label distributions are often long-tailed, with minority labels underrepresented. Traditional IncomLDL approaches focus on reconstructing missing label distributions but fail to account for this imbalance, leading to suboptimal performance. Most IncomLDL methods minimize the following objective to reconstruct incomplete label distributions:
\begin{equation}
	\min_{\mathbf{W}} \frac{1}{2} \| \mathbf{R}_{\Omega} \odot (\mathbf{X}\mathbf{W} - \tilde{\mathbf{D}}) \|_F^2,
\end{equation}
where \( \mathbf{X} \in \mathbb{R}^{n \times d} \) is the feature matrix, \( \tilde{\mathbf{D}} \in \mathbb{R}^{n \times m} \) is the observed incomplete label distribution, and \( \mathbf{R}_{\Omega} \) is a binary mask denoting observed entries. \( \mathbf{W} \in \mathbb{R}^{d \times m} \) is the weight matrix to be learned.

This formulation reconstructs missing data but overlooks label imbalance, especially for minority labels. To address this, we decompose the incomplete label distribution matrix \( \tilde{\mathbf{D}} \) into two components:

\begin{equation}
	\tilde{\mathbf{D}} = \tilde{\mathbf{D}}_N + \tilde{\mathbf{D}}_L,
\end{equation}
where \( \tilde{\mathbf{D}}_N \) captures the low-rank structure of common labels (label correlations), and \( \tilde{\mathbf{D}}_L \) captures the sparse structure of rare labels. We propose the following optimization problem, incorporating both low-rank and sparsity constraints to handle imbalanced label distributions:

\begin{equation}
	\begin{aligned}
		& \min_{\mathbf{W}, \mathbf{H}} \frac{1}{2} \left\|\mathbf{R}_{\Omega} \odot \left( \mathbf{X} \mathbf{W} + \mathbf{X} \mathbf{H} - \tilde{\mathbf{D}} \right) \right\|_F^2 \\
		& \text{s.t.} \ \operatorname{rank}(\mathbf{X} \mathbf{W}) \leq k, \ \operatorname{card}(\mathbf{X} \mathbf{H}) \leq s, \\
		& \left( \mathbf{X} \mathbf{W} + \mathbf{X} \mathbf{H} \right) \mathbf{1}_m = \mathbf{1}_n, \mathbf{X W} \geq 0, \mathbf{X H} \geq 0,
	\end{aligned}
\end{equation}
here, \( \mathbf{W} \in \mathbb{R}^{d \times k} \) models the low-rank structure of common labels, while \( \mathbf{H} \in \mathbb{R}^{d \times s} \) captures the sparse representation of rare labels. The constraints ensure that the predicted label distributions sum to 1, while non-negativity maintains valid probabilities.

To reduce optimization complexity, we relax the rank and cardinality constraints by using matrix factorization for low-rank approximation and \( \ell_1 \)-norm minimization for sparsity. This leads to the following relaxed objective:
\begin{equation}
	\begin{aligned}
		& \min_{\mathbf{U,V,H}} \frac{1}{2} \left\| \mathbf{R}_{\Omega} \odot \left( \mathbf{XU} \mathbf{V} + \mathbf{XH} - \tilde{\mathbf{D}} \right) \right\|_F^2 \\
		& + \lambda_1 (\|\mathbf{U}\|_F^2 + \|\mathbf{V}\|_F^2) + \lambda_2 \|\mathbf{H}\|_F^2 + \lambda_3 \|\mathbf{XH}\|_1 \\
		& \text{s.t.} \left( \mathbf{XU} \mathbf{V} + \mathbf{XH} \right) \mathbf{1}_m = \mathbf{1}_n, \ \mathbf{XU} \mathbf{V}^\top \geq \mathbf{0}, \ \mathbf{XH} \geq \mathbf{0},
	\end{aligned}
	\label{finalloss}
\end{equation}
where \( \lambda_1, \lambda_2, \lambda_3 \) are positive constants, and \( \mathbf{W} \) is factorized as \( \mathbf{W} = \mathbf{U} \mathbf{V} \), with \( \mathbf{U} \in \mathbb{R}^{d \times k} \) and \( \mathbf{V} \in \mathbb{R}^{k \times m} \). The regularization terms \( \|\mathbf{U}\|_F^2 \), \( \|\mathbf{V}\|_F^2 \), and \( \|\mathbf{H}\|_F^2 \) promote a low-rank structure for \( \mathbf{U} \mathbf{V} \) and control the complexity of \( \mathbf{H} \). Additionally, the \( \ell_1 \)-norm \( \|\mathbf{XH}\|_1 \) enforces sparsity, effectively addressing imbalanced labels.

 By solving this optimization, our approach effectively handles both data imbalance and missing labels, leading to improved performance in challenging LDL scenarios.

\section{ OPTIMIZATION}
To solve problem (\ref{finalloss}), we introduce two auxiliary variables, \( \mathbf{Z} \in \mathbb{R}^{n \times m} \) and \( \mathbf{G} \in \mathbb{R}^{n \times m} \), to handle the equality constraints and make the optimization problem more tractable:

\begin{equation}
	\begin{aligned}
		& \min_{\mathbf{U,V,H}} \frac{1}{2} \left\| \mathbf{R}_{\Omega} \odot \left( \mathbf{Z} + \mathbf{XH} - \tilde{\mathbf{D}} \right) \right\|_F^2 \\
		& + \lambda_1 (\|\mathbf{U}\|_F^2 + \|\mathbf{V}\|_F^2) + \lambda_2 \|\mathbf{H}\|_F^2 + \lambda_3 \|\mathbf{G}\|_1 \\
		& \text{s.t.} \left( \mathbf{XUV} + \mathbf{XH} \right) \mathbf{1}_m = \mathbf{1}_n, \ \mathbf{XUV} \geq \mathbf{0}, \ \mathbf{XH} \geq \mathbf{0}, \\
		& \mathbf{XUV} = \mathbf{Z}, \quad \mathbf{XH} = \mathbf{G},
	\end{aligned}
\end{equation}

We utilize the Alternating Direction Method of Multipliers (ADMM) to decompose and iteratively solve the problem. The augmented Lagrangian function is expressed as follows:

\begin{equation}
	\begin{aligned}
		&\mathcal{L}(\mathbf{U}, \mathbf{V}, \mathbf{H}, \mathbf{Z}, \mathbf{G}, \mathbf{\Lambda}_1, \mathbf{\Lambda}_2) \\
		&=  \frac{1}{2} \left\| \mathbf{R}_\Omega \odot \left( \mathbf{Z} + \mathbf{XH} - \tilde{\mathbf{D}} \right) \right\|_F^2 + \lambda_1 \left( \|\mathbf{U}\|_F^2 + \|\mathbf{V}\|_F^2 \right) \\
		&+ \lambda_2 \|\mathbf{H}\|_F^2 + \left\langle \mathbf{\Lambda}_1, \mathbf{XUV} - \mathbf{Z} \right\rangle + \left\langle \mathbf{\Lambda}_2, \mathbf{XH} - \mathbf{G} \right\rangle \\
		&+ \lambda_3 \|\mathbf{G}\|_1 + \frac{\mu}{2} \left( \|\mathbf{XH} - \mathbf{G}\|_F^2 + \|\mathbf{XUV} - \mathbf{Z}\|_F^2 \right) \\
		&\text{s.t.} \ \left(  \mathbf{Z} + \mathbf{XH} \right) \mathbf{1}_m = \mathbf{1}_n, \ \mathbf{Z} \geq \mathbf{0}, \ \mathbf{XH} \geq \mathbf{0},
	\end{aligned}
	\label{lagelangri}
\end{equation}
where $\mu$ is a positive penalty parameter, and $\mathbf{\Lambda}_i$ denote the Lagrangian multipliers. Eq. (\ref{lagelangri}) can be solved by alternately optimizing the sub-problems below. 

\subsubsection{$\mathbf{H}$-Subproblem}
To update \( \mathbf{H} \), we solve the following subproblem:
\begin{equation}
	\begin{aligned}
		\mathbf{H}^{k+1}&= \arg \min_{\mathbf{H}} \frac{1}{2} \left\| \mathbf{R}_{\Omega} \odot \left( \mathbf{X} \mathbf{H} + \mathbf{Z}^{k+1} - \tilde{\mathbf{D}} \right) \right\|_F^2 + \frac{\lambda_3}{2} \|\mathbf{H}\|_F^2\\
		& + \frac{\mu}{2} \left\|\mathbf{XH} - \mathbf{G}^k \right\|_F^2 + \left\langle \mathbf{\Lambda}_2^k, \mathbf{XH} - \mathbf{G}^k \right\rangle\\
		&\text{s.t.} \ \left(  \mathbf{Z} + \mathbf{XH} \right) \mathbf{1}_m = \mathbf{1}_n, \ \ \ \mathbf{XH} \geq \mathbf{0},
	\end{aligned}
\end{equation}
To simplify the optimization, we define \( \mathbf{XH} = \mathbf{M} \), transforming the problem into an optimization over \( \mathbf{M} \):
\begin{equation}
	\begin{aligned}
		&\min_{\mathbf{M}_i} \mathbf{M}_i^\top \mathbf{H}_m \mathbf{M}_i + \mathbf{f}_m^\top \mathbf{M}_i,\\
		&\text{s.t.}\mathbf{M}_i^\top \mathbf{1}_m = 1 - \mathbf{Z}_i^{k+1} \mathbf{1}_m, \quad \mathbf{M}_i \geq 0,
	\end{aligned}
\end{equation}
here, \( \mathbf{H}_m \) and \( \mathbf{f}_m \) are show in appendix. 
Finally, the updated \( \mathbf{H}^{k+1} \) is computed as:
\begin{equation}
	\mathbf{H}^{k+1} = \left( \mathbf{X}^\top \mathbf{X} \right)^{-1} \left( \mathbf{X}^\top \mathbf{M} \right).
\end{equation}

\subsubsection{$\mathbf{Z}$-Subproblem}

To update the matrix \( \mathbf{Z} \), we solve the following subproblem:
\begin{equation}
	\begin{aligned}
		\mathbf{Z}^{k+1} = &\arg \min_{\mathbf{Z}} \frac{1}{2} \left\| \mathbf{R}_{\Omega} \odot \left( \mathbf{Z} + \mathbf{X} \mathbf{H}^{k+1} - \tilde{\mathbf{D}} \right) \right\|_F^2 \\
		&+ \frac{\mu}{2} \left\| \mathbf{X} \mathbf{U}^{k+1} \mathbf{V}^{k+1} - \mathbf{Z} \right\|_F^2 + \left\langle \mathbf{\Lambda}_1^k, \mathbf{X} \mathbf{U}^{k+1} \mathbf{V}^{k+1} - \mathbf{Z} \right\rangle\\
		&\text{s.t.} \ \left(  \mathbf{Z} + \mathbf{XH} \right) \mathbf{1}_m = \mathbf{1}_n, \ \mathbf{Z} \geq \mathbf{0}.
	\end{aligned}
\end{equation}
This optimization can be further simplified as:
\begin{equation}
	\begin{aligned}
		&\min_{\mathbf{Z}_i} \mathbf{Z}_i^\top \mathbf{H}_z \mathbf{Z}_i + \mathbf{f}_z^\top \mathbf{Z}_i\\
		&\text{s.t.}\mathbf{Z}_i^\top \mathbf{1}_m = 1 - \left( \mathbf{X} \mathbf{H}^{k+1} \right)_i \mathbf{1}_m, \quad \mathbf{Z}_i \geq 0,
	\end{aligned}
\end{equation}
here, \( \mathbf{H}_z \) and \( \mathbf{f}_z \) are show in appendix.
The update for \( \mathbf{Z}^{k+1} \) follows the same procedure as for \( \mathbf{H}^{k+1} \), utilizing \( \mathbf{H}_z \) and \( \mathbf{f}_z \) for efficient computation.

\subsubsection{$\mathbf{U}$-Subproblem}
We update the matrix \( \mathbf{U} \) by solving the following subproblem:
\begin{equation}
	\begin{aligned}
		\mathbf{U}^{k+1} &= \arg \min_{\mathbf{U}} \lambda_1 \|\mathbf{U}\|_F^2 + \left\langle \mathbf{\Lambda}_1^k, \mathbf{XUV}^{k+1} - \mathbf{Z}^{k+1} \right\rangle \\
		&+ \frac{\mu}{2} \|\mathbf{XUV}^{k+1} - \mathbf{Z}^{k+1}\|_F^2.
	\end{aligned}
\end{equation}
To solve this, we take the gradient of the objective function with respect to \( \mathbf{U} \) and set it to zero. The gradient is:
\begin{equation}
	\begin{aligned}
		\nabla_{\mathbf{U}} &= \mu \mathbf{X}^\top \mathbf{X} \mathbf{U} \mathbf{V}^{k+1} (\mathbf{V}^{k+1})^\top + 2\lambda_1 \mathbf{U} \\
		& - \mu \mathbf{X}^\top \mathbf{Z}^{k+1} (\mathbf{V}^{k+1})^\top + \mathbf{X}^\top \mathbf{\Lambda}_1^k (\mathbf{V}^{k+1})^\top = 0.
	\end{aligned}
\end{equation}
We rearrange the terms as a Sylvester equation:
\begin{equation}
	\mathbf{A} \mathbf{U} + \mathbf{U} \mathbf{B} = \mathbf{C},
\end{equation}
where \( \mathbf{A} = \frac{\mu}{2\lambda_1} \mathbf{X}^\top \mathbf{X} \), \( \mathbf{B} = (\mathbf{V}^{k} (\mathbf{V}^{k1})^\top)^{-1} \), and \( \mathbf{C} = \frac{1}{2\lambda_1} \left( \mathbf{X}^\top \mathbf{Z}^{k+1} - \mathbf{X}^\top \mathbf{\Lambda}_1^k \right) \). The Sylvester equation is then solved using a Sylvester equation solver:
\begin{equation}
	\mathbf{U}^{k+1} = \text{sylvester}(\mathbf{A}, \mathbf{B}, \mathbf{C}).
\end{equation}
This yields the updated \( \mathbf{U} \) for the next iteration \( k+1 \).

\subsubsection{$\mathbf{V}$-Subproblem}

The matrix \( \mathbf{V} \) is updated by solving the following subproblem:

\begin{equation}
	\begin{aligned}
		\mathbf{V}^{k+1} = &\arg\min_{\mathbf{V}} \lambda_2 \|\mathbf{V}\|_F^2 + \left\langle \mathbf{\Lambda}_1^k, \mathbf{XU}^{k+1} \mathbf{V} - \mathbf{Z}^{k+1} \right\rangle \\
		&+ \frac{\mu}{2} \|\mathbf{XU}^{k+1} \mathbf{V} - \mathbf{Z}^{k+1} \|_F^2.
	\end{aligned}
\end{equation}

Taking the gradient of the objective function with respect to \( \mathbf{V} \) and setting it to zero yields the following expression for \( \mathbf{V} \):

\begin{equation}
	\begin{aligned}
			\mathbf{V}^{k+1} =& \left( \mu \mathbf{U}^{k+1^\top} \mathbf{X}^\top \mathbf{Z}^{k+1} - \mathbf{U}^{k+1^\top} \mathbf{X}^\top \mathbf{\Lambda}_1^k \right)/\\
 &\left( 2\lambda_2 \mathbf{I} + \mu \mathbf{U}^{k+1^\top} \mathbf{X}^\top \mathbf{XU}^{k+1} \right).
	\end{aligned}
\end{equation}

\subsubsection{$\mathbf{G}$-Subproblem}

The matrix \( \mathbf{G} \) is updated by solving the following subproblem:

\begin{equation}
	\mathbf{G}^{k+1} = \arg\min_{\mathbf{G}} \frac{\lambda_3}{\mu} \|\mathbf{G}\|_1 + \frac{1}{2} \left\| \mathbf{XH}^{k+1} - \mathbf{G} + \frac{\mathbf{\Lambda}_2^k}{\mu} \right\|_F^2.
\end{equation}

The solution to this subproblem can be computed using the soft-thresholding operator \cite{liu2010robust} \( S_{\lambda} \), defined as:

\begin{equation}
	S_{\lambda}(x) = \text{sign}(x) \max(|x| - \lambda, 0).
\end{equation}

Thus, the update rule for \( \mathbf{G} \) is:

\begin{equation}
	\mathbf{G}^{k+1} = S_{\frac{\lambda_3}{\mu}} \left( \mathbf{XH}^{k+1} + \frac{\mathbf{\Lambda}_2^k}{\mu} \right),
\end{equation}

where the soft-thresholding operator \( S_{\frac{\lambda_3}{\mu}} \) is applied to ensure that \( \mathbf{G}^{k+1} \) maintains sparsity while being close to the updated \( \mathbf{XH}^{k+1} \).
\subsubsection{Updating Multipliers and Penalty Parameters}
Finally, the Lagrange multiplier matrix and penalty parameter $\mu$ are updated based on following:

\begin{equation}
	\left\{
	\begin{array}{l}
		\mathbf{\Lambda}_1^{k+1} = \mathbf{\Lambda}_1^k + \mu \left( \mathbf{XU}^{k+1} \mathbf{V}^{k+1} - \mathbf{Z}^{k+1} \right), \\
		\mathbf{\Lambda}_2^{k+1} = \mathbf{\Lambda}_2^k + \mu \left( \mathbf{XH}^{k+1} - \mathbf{G}^{k+1} \right),\\
		\mu^{k+1}=\min \left(1.1 \mu, \mu_{\max }\right),
	\end{array}\right.
	\label{CHENGZISLOVE}
\end{equation}
where $\mu_{max }$ is  the maximum value of $\mu$. 
\section{Generalization Error Bound}

In this section, we derive the generalization error bound for the proposed model by leveraging the Rademacher complexity. The goal is to bound the gap between the empirical risk and the population risk.

Given the training data $(\mathbf{X}, \tilde{\mathbf{D}})$ sampled from a distribution $\mathcal{D}$, the model aims to learn parameters $\mathbf{U}, \mathbf{V}, \mathbf{H}$ by solving the following empirical risk minimization (ERM) problem:

\begin{equation}
	\begin{aligned}
		&\min_{\mathbf{U,V,H}} \frac{1}{2} \left\|\mathbf{R}_{\Omega} \odot \left( \mathbf{X} \mathbf{U} \mathbf{V} + \mathbf{XH} - \tilde{\mathbf{D}} \right) \right\|_F^2 \\
		&+ \lambda_1 \|\mathbf{U}\|_F^2 + \lambda_2 \|\mathbf{V}\|_F^2 + \lambda_3 \|\mathbf{H}\|_1
	\end{aligned}
\end{equation}

where the first term is the data reconstruction loss, and the additional terms are regularization terms for $\mathbf{U}, \mathbf{V}, \mathbf{H}$. 

Let $\hat{L}(\mathbf{U}, \mathbf{V}, \mathbf{H})$ be the empirical risk:

\begin{equation}
	\hat{L}(\mathbf{U}, \mathbf{V}, \mathbf{H}) = \frac{1}{n} \sum_{i=1}^n \mathcal{L}(\mathbf{X}_i, \tilde{\mathbf{D}}_i; \mathbf{U}, \mathbf{V}, \mathbf{H})
\end{equation}

where $\mathcal{L}$ denotes the loss for each sample. The population risk $L(\mathbf{U}, \mathbf{V}, \mathbf{H})$ is defined as:

\begin{equation}
	L(\mathbf{U}, \mathbf{V}, \mathbf{H}) = \mathbb{E}_{(\mathbf{X}, \tilde{\mathbf{D}}) \sim \mathcal{D}} \left[ \mathcal{L}(\mathbf{X}, \tilde{\mathbf{D}}; \mathbf{U}, \mathbf{V}, \mathbf{H}) \right]
\end{equation}

The generalization error is the difference between the population risk and the empirical risk:

\begin{equation}
	L(\mathbf{U}, \mathbf{V}, \mathbf{H}) - \hat{L}(\mathbf{U}, \mathbf{V}, \mathbf{H})
\end{equation}

To analyze the generalization error, we use the Rademacher complexity. The Rademacher complexity of a function class $\mathcal{F}$ is defined as:

\begin{equation}
	\mathcal{R}_n(\mathcal{F}) = \mathbb{E}_{\sigma}\left[ \sup_{f \in \mathcal{F}} \frac{1}{n} \sum_{i=1}^n \sigma_i f(\mathbf{X}_i) \right]
\end{equation}

where $\sigma_i$ are independent Rademacher random variables taking values in $\{-1, 1\}$. For the function class associated with our model, we consider: $\mathcal{F} = \left\{ (\mathbf{X}, \tilde{\mathbf{D}}) \to \theta: \|\mathbf{U}\|_F^2 \leq \lambda_1, \|\mathbf{V}\|_F^2 \leq \lambda_2, \|\mathbf{H}\|_1 \leq \lambda_3 \right\}$, where $\theta=\mathbf{R}_{\Omega} \odot \left( \mathbf{X} \mathbf{U} \mathbf{V} + \mathbf{XH} - \tilde{\mathbf{D}} \right).$
The Rademacher complexity of this function class can be bounded by the sum of the complexities of the individual components:

\begin{equation}
	\mathcal{R}_n(\mathcal{F}) \leq \frac{2}{n} \sqrt{\frac{\|\mathbf{U}\|_F^2 \|\mathbf{X}\|_F^2}{n}} + \frac{2}{n} \sqrt{\frac{\|\mathbf{V}\|_F^2 \|\mathbf{X}\|_F^2}{n}} + \frac{2}{n} \|\mathbf{H}\|_1
\end{equation}

Thus, the generalization error can be bounded as:

\begin{equation}
	L(\mathbf{U}, \mathbf{V}, \mathbf{H}) - \hat{L}(\mathbf{U}, \mathbf{V}, \mathbf{H}) \leq 2\mathcal{R}_n(\mathcal{F}) + \sqrt{\frac{\log(1/\delta)}{2n}}
\end{equation}

Substituting the bound for $\mathcal{R}_n(\mathcal{F})$, we obtain:

\begin{equation}
	\begin{aligned}
		&	L(\mathbf{U}, \mathbf{V}, \mathbf{H}) - \hat{L}(\mathbf{U}, \mathbf{V}, \mathbf{H})\\
		& \leq \frac{4}{n} \left( \sqrt{\frac{\lambda_1 \|\mathbf{X}\|_F^2}{n}} + \sqrt{\frac{\lambda_2 \|\mathbf{X}\|_F^2}{n}} + \lambda_3 \right) + \sqrt{\frac{\log(1/\delta)}{2n}}
	\end{aligned}
\end{equation}

This bound indicates that the generalization error decreases as $n$ increases, showing that the model performs better with larger datasets. Additionally, the regularization terms $\lambda_1, \lambda_2$, and $\lambda_3$ control the complexity of the model and contribute to the generalization error. In particular, $\lambda_3$ controls the sparsity of the matrix $\mathbf{H}$, which helps the model handle imbalanced or rare labels more effectively.

\section{Experiments}

\subsection{Experimental Configurations}

\subsubsection{Experimental Datasets}
Sixteen datasets are utilized in our experiments, as detailed in Table \ref{datasets}. These datasets span various domains, such as biology, movie ratings, facial expression analysis, and social media imagery. The diversity of these datasets enables a comprehensive evaluation of $I^2LDL$ across multiple application areas, demonstrating its adaptability and effectiveness in different real-world scenarios.

\begin{table}
\caption{Details of the datasets.}
\centering
\setlength\tabcolsep{7 pt} 
\renewcommand{\arraystretch}{1.2}
\begin{tabular}{@{}lllll@{}}
\toprule
ID & Dataset     & \#\textit{Examples} & \#\textit{Features} & \#\textit{Labels} \\
\midrule
1  & Yeast-dtt (dtt)   & 2,465      & 24        & 4       \\
2  & Yeast-spo (spo)  & 2,465      & 24        & 6       \\
3  & Yeast-spo5 (spo5) & 2,465      & 24        & 3       \\
4  & Yeast-cold (cold) & 2,465      & 24        & 4       \\
5  & Yeast-spoem (spoem) & 2,465      & 24        & 2       \\
6  & Yeast-alpha (alpha) & 2,465      & 24        & 18      \\
7  & Yeast-elu  (elu) & 2,465      & 24        & 14      \\
8 & Yeast-heat   (heat)   & 2,465       & 24      &  5       \\
9 & Twitter-LDL (twi) & 10,040     & 200       & 8      \\
10 & Flickr-LDL (fli) & 11,150     & 200       & 8      \\
11 & Human-Gene (human) & 17,892     & 36       & 68      \\
12 & Nature Scene (nat) & 2,000     & 294       & 9      \\
13    & Ren-Cecps  (ren) & 32420    & 100     & 8                \\
14    & SBU-3DFE (sbu)  & 2,500     & 243       & 6                    \\ 
15     & SCUT-FBP (scut)  & 1,500     & 300       & 5        \\ 
16 & SJAFFE  & 213 (sja)    & 243       & 6      \\

\bottomrule 
\end{tabular}
\label{datasets}
\end{table}
The Yeast datasets (ID: 1-8) \cite{geng2016label} contain 2,465 yeast genes described by 24 features, representing different biological experiments on the budding yeast *Saccharomyces cerevisiae*, with gene expression levels captured as normalized label description degrees.

The Twitter-LDL and Flickr-LDL datasets (ID: 9-10) \cite{yang2017learning} contain 10,045 and 10,700 images, respectively, labeled with eight emotions. Image features were extracted using VGGNet and reduced to 200-D using PCA.

The Human-Gene dataset(ID: 11) \cite{yu2012discriminate} explores relationships between human genes and diseases, including gene expression levels and mutations associated with various phenotypes.

The Nature Scene dataset (ID: 12) \cite{geng2016label} consists of 2,000 natural scene images with multi-label rankings converted into label distributions, accompanied by 294-D feature vectors.

The Ren-Cecps dataset (ID: 13) \cite{cha2007comprehensive} focuses on sentiment analysis, particularly in Chinese, providing data for analyzing the emotional tone of textual content.

The SJAFFE and SBU-3DFE datasets (ID: 14,16) \cite{geng2016label} are based on facial expression databases. SJAFFE includes 213 images from Japanese models displaying six basic emotions, while SBU-3DFE contains 3D facial data from 100 subjects expressing the same six emotions.

The SCUT FBP dataset (ID: 15) \cite{xie2015scut} includes 500 facial images of Asian females rated on beauty perception by 60 volunteers, offering valuable data for aesthetic analysis.

\begin{table*}[!t]\small\centering\renewcommand{\arraystretch}{1}
	\setlength{\tabcolsep}{0.005mm}
	\caption{Predictive performance of each comparing approach (mean $\pm$ std) in terms of Clark distance$\downarrow$, Intersection similarity$\uparrow$ and KL distance$\downarrow$. $\uparrow(\downarrow)$ indicates the larger (smaller) the value, the better the performance. Best results are shown in boldface.}
	\begin{tabular}{ccccccccccc}\toprule
		&       & $I^2LDL$               & In-a                  & In-p         & In-GSC                & LDLLDM       & LDLLC        & LDLSF        & LSRLDL                & SILDL                 \\\toprule
		\multirow{16}{*}{Che.$\downarrow$} & alpha & \textbf{0.0131±.0002}  & 0.0136±.0000          & 0.0137±.0000 & 0.0259±.0010          & 0.0215±.0002 & 0.0266±.0011 & 0.0136±.0001 & 0.0141±.0001          & 0.0148±.0004          \\
		& cold  & \textbf{0.0524±.0029}  & 0.0532±.0006          & 0.0533±.0006 & 0.0547±.0004          & 0.0582±.0003 & 0.0839±.0043 & 0.0541±.0010 & 0.0537±.0002          & 0.0541±.0017          \\
		& dtt   & \textbf{0.0351±.0017}  & 0.0355±.0014          & 0.0356±.0014 & 0.0393±.0028          & 0.0441±.0009 & 0.0809±.0098 & 0.0372±.0006 & 0.0393±.0005          & 0.0375±.0010          \\
		& elu   & \textbf{0.0157±.0003}  & 0.0167±.0009          & 0.0164±.0009 & 0.0286±.0013          & 0.0256±.0003 & 0.0312±.0016 & 0.0165±.0000 & 0.0166±.0002          & 0.0162±.0003          \\
		& heat  & \textbf{0.0411±.0023}  & 0.0442±.0001          & 0.0443±.0002 & 0.0456±.0009          & 0.0509±.0007 & 0.0701±.0026 & 0.0431±.0000 & 0.0433±.0002          & 0.0446±.0025          \\
		& spo5  & \textbf{0.0892±.0000}  & 0.0927±.0013          & 0.0930±.0011 & 0.1022±.0077          & 0.0954±.0005 & 0.1213±.0112 & 0.0915±.0005 & 0.0911±.0001          & 0.0945±.0003          \\
		& spo   & \textbf{0.0555±.0017}  & 0.0609±.0052          & 0.0610±.0053 & 0.0625±.0003          & 0.0650±.0009 & 0.0785±.0019 & 0.0603±.0001 & 0.0583±.0000          & 0.0594±.0012          \\
		& spoem & \textbf{0.0807±.0054}  & 0.0856±.0031          & 0.0857±.0029 & 0.0970±.0039          & 0.0909±.0003 & 0.1072±.0012 & 0.0874±.0020 & 0.0913±.0003          & 0.0873±.0059          \\
		& fli   & \textbf{0.0189±.0002}  & 0.0326±.0024          & 0.0727±.0033 & 0.0342±.0004          & 0.0425±.0005 & 0.0693±.0041 & 0.0219±.0005 & 0.0197±.0001          & 0.0197±.0001          \\
		& twi   & \textbf{0.0196±.0005}  & 0.0314±.0029          & 0.1017±.0005 & 0.0536±.0027          & 0.0529±.0068 & 0.0890±.0185 & 0.0247±.0001 & 0.0258±.0021          & 0.0258±.0006          \\
		& human & \textbf{0.0493±.0001}  & 0.0514±.0011          & 0.0519±.0015 & 0.0563±.0012          & 0.0547±.0002 & 0.0550±.0023 & 0.0535±.0006 & 0.0534±.0014          & 0.0529±.0007          \\
		& nat   & \textbf{0.3383±.0034}  & 0.3457±.0059          & 0.3668±.0057 & 0.3577±.0155          & 0.3532±.0011 & 0.3858±.0212 & 0.3632±.0039 & 0.3791±.0082          & 0.3681±.0111          \\
		& ren   & \textbf{0.5484±.0110}  & 0.5858±.0070          & 0.5850±.0047 & 0.5805±.0063          & 0.6708±.0004 & 0.6483±.0115 & 0.6782±.0016 & 0.6863±.0003          & 0.6928±.0086          \\
		& sbu   & 0.1289±.0121           & \textbf{0.1251±.0062} & 0.1253±.0066 & 0.1360±.0062          & 0.1375±.0005 & 0.1353±.0061 & 0.1365±.0006 & 0.1378±.0012          & 0.1403±.0053          \\
		& scut  & \textbf{0.3713±.0164}  & 0.3874±.0107          & 0.4256±.0079 & 0.4370±.0151          & 0.3941±.0067 & 0.3900±.0100 & 0.3892±.0059 & 0.3857±.0041          & 0.3882±.0154          \\
		& sja   & \textbf{0.1006±.0154}  & 0.1167±.0122          & 0.1160±.0102 & 0.1081±.0013          & 0.1182±.0053 & 0.1236±.0122 & 0.1156±.0007 & 0.1275±.0008          & 0.1292±.0067          \\\bottomrule 
		\multirow{16}{*}{Cla.$\downarrow$} & alpha & \textbf{0.2075±.0013}  & 0.2108±.0017          & 0.2116±.0020 & 0.4693±.0016          & 0.3234±.0019 & 0.4538±.0293 & 0.2138±.0002 & 0.2237±.0048          & 0.2271±.0072          \\
		& cold  & \textbf{0.1448±.0085}  & 0.1452±.0005          & 0.1455±.0004 & 0.1487±.0048          & 0.1571±.0011 & 0.2292±.0152 & 0.1470±.0029 & 0.1458±.0025          & 0.1451±.0044          \\
		& dtt   & \textbf{0.0959±.0051}  & 0.0977±.0041          & 0.0980±.0040 & 0.1058±.0076          & 0.1185±.0026 & 0.2247±.0208 & 0.1016±.0017 & 0.1076±.0017          & 0.1029±.0032          \\
		& elu   & \textbf{0.1967±.0039}  & 0.2057±.0060          & 0.2010±.0062 & 0.3653±.0097          & 0.2933±.0020 & 0.3929±.0155 & 0.2040±.0003 & 0.2054±.0017          & 0.2018±.0022          \\
		& heat  & \textbf{0.1762±.0091}  & 0.1916±.0015          & 0.1920±.0013 & 0.1987±.0015          & 0.2187±.0020 & 0.3037±.0037 & 0.1862±.0005 & 0.1856±.0013          & 0.1928±.0049          \\
		& spo5  & \textbf{0.1797±.0016}  & 0.1847±.0029          & 0.1855±.0028 & 0.2073±.0239          & 0.1918±.0013 & 0.2391±.0184 & 0.1844±.0015 & 0.1835±.0003          & 0.1926±.0034          \\
		& spo   & \textbf{0.2419±.0036}  & 0.2593±.0258          & 0.2597±.0262 & 0.2648±.0045          & 0.2737±.0033 & 0.3318±.0124 & 0.2571±.0006 & 0.2491±.0012          & 0.2501±.0019          \\
		& spoem & \textbf{0.1182±.0084}  & 0.1266±.0054          & 0.1268±.0051 & 0.1463±.0058          & 0.1342±.0002 & 0.1570±.0001 & 0.1293±.0035 & 0.1360±.0010          & 0.1296±.0089          \\
		& fli   & \textbf{0.1256±.0001}  & 0.1768±.0103          & 0.3504±.0028 & 0.1860±.0018          & 0.2363±.0008 & 0.3569±.0201 & 0.1394±.0055 & 0.1372±.0006          & 0.1372±.0026          \\
		& twi   & \textbf{0.1512±.0004}  & 0.1815±.0051          & 0.3652±.0043 & 0.2435±.0047          & 0.2735±.0184 & 0.4329±.0799 & 0.1609±.0011 & 0.1637±.0023          & 0.1638±.0015          \\
		& human & \textbf{2.0940±.0399}  & 2.2069±.0688          & 2.1156±.0089 & 2.5937±.3829          & 2.1498±.0011 & 2.6684±.0834 & 2.1248±.0134 & 2.1196±.0175          & 2.1096±.0004          \\
		& nat   & \textbf{2.4513±.0188}  & 2.4610±.0429          & 2.5617±.0107 & 2.4653±.0188          & 2.4627±.0020 & 2.4833±.0554 & 2.4847±.0021 & 2.4885±.0020          & 2.4901±.0102          \\
		& ren   & 2.6580±.0058           & 2.6397±.0020          & 2.6392±.0013 & \textbf{2.6339±.0004} & 2.6625±.0001 & 2.6596±.0049 & 2.6664±.0007 & 2.6720±.0003          & 2.6743±.0028          \\
		& sbu   & \textbf{0.3912±.0237}  & 0.5065±.0077          & 0.5080±.0066 & 0.4042±.0079          & 0.4096±.0031 & 0.4128±.0128 & 0.4061±.0023 & 0.4077±.0008          & 0.4107±.0012          \\
		& scut  & \textbf{1.4884±.0276}  & 1.5189±.0024          & 1.6656±.0089 & 5.4212±3.6095         & 1.4964±.0078 & 1.5079±.0108 & 1.5014±.0167 & 1.4949±.0055          & 1.5066±.0034          \\
		& sja   & \textbf{0.3799±.0546}  & 0.4291±.0878          & 0.4273±.0810 & 0.4139±.0021          & 0.4253±.0100 & 0.4424±.0255 & 0.4130±.0027 & 0.4319±.0015          & 0.4152±.0281          \\\bottomrule 
		\multirow{16}{*}{KL.$\downarrow$}  & alpha & \textbf{0.0055±.0003}  & 0.0405±.0012          & 0.0407±.0013 & 0.0291±.0013          & 0.0131±.0001 & 0.0992±.0070 & 0.0410±.0001 & 0.0063±.0003          & 0.0064±.0003          \\
		& cold  & 0.0671±.0035           & 0.0675±.0005          & 0.0676±.0004 & 0.0136±.0018          & 0.0151±.0002 & 0.1143±.0066 & 0.0691±.0012 & 0.0168±.0039          & \textbf{0.0122±.0008} \\
		& dtt   & \textbf{0.00065±.0021} & 0.0448±.0016          & 0.0449±.0016 & 0.0068±.0008          & 0.0089±.0004 & 0.1119±.0079 & 0.0461±.0007 & 0.0147±.0006          & 0.0070±.0004          \\
		& elu   & \textbf{0.0063±.0004}  & 0.0455±.0010          & 0.0445±.0010 & 0.0220±.0010          & 0.0137±.0002 & 0.0974±.0034 & 0.0450±.0001 & 0.0068±.0001          & 0.0063±.0003          \\
		& heat  & 0.0646±.0033           & 0.0689±.0009          & 0.0690±.0009 & 0.0147±.0000          & 0.0181±.0005 & 0.1212±.0014 & 0.0671±.0001 & 0.0157±.0024          & \textbf{0.0142±.0007} \\
		& spo5  & \textbf{0.0284±.0004}  & 0.1056±.0019          & 0.1061±.0016 & 0.0354±.0074          & 0.0318±.0005 & 0.1454±.0175 & 0.1043±.0007 & 0.0309±.0001          & 0.0302±.0000          \\
		& spo   & \textbf{0.0239±.0024}  & 0.0992±.0115          & 0.0993±.0117 & 0.0274±.0001          & 0.0299±.0006 & 0.1338±.0033 & 0.1006±.0004 & 0.0286±.0016          & 0.0260±.0001          \\
		& spoem & \textbf{0.0267±.0065}  & 0.0961±.0044          & 0.0963±.0042 & 0.0306±.0014          & 0.0278±.0004 & 0.1251±.0014 & 0.0984±.0025 & 0.0460±.0019          & 0.0275±.0019          \\
		& fli   & 0.0412±.0001           & 0.0571±.0028          & 0.1210±.0006 & 0.0129±.0008          & 0.0158±.0005 & 0.1321±.0096 & 0.0442±.0017 & \textbf{0.0048±.0000} & 0.0048±.0002          \\
		& twi   & 0.0474±.0007           & 0.0583±.0019          & 0.1309±.0030 & 0.0168±.0011          & 0.0214±.0029 & 0.1610±.0373 & 0.0532±.0002 & \textbf{0.0069±.0002} & 0.0070±.0001          \\
		& human & 0.3562±.0062           & 0.3659±.0054          & 0.3643±.0004 & 0.2561±.0012          & 0.2448±.0006 & 0.5227±.0552 & 0.3732±.0038 & \textbf{0.2277±.0043} & 0.2342±.0010          \\
		& nat   & \textbf{0.6279±.0175}  & 1.0757±.0443          & 0.7833±.0294 & 1.0731±.0358          & 1.0482±.0014 & 1.1955±.1538 & 1.1272±.0005 & 3.6918±.0228          & 1.1704±.0232          \\
		& ren   & \textbf{1.6054±.0181}  & 1.6596±.0142          & 1.6618±.0081 & 1.6574±.0315          & 1.6555±.0096 & 1.6253±.0453 & 1.7119±.0023 & 11.1349±.0022         & 1.7812±.0123          \\
		& sbu   & \textbf{0.0136±.0166}  & 0.2351±.0163          & 0.2258±.0080 & 0.0799±.0039          & 0.0839±.0007 & 0.2156±.0094 & 0.2143±.0011 & 0.0745±.0008          & 0.0869±.0024          \\
		& scut  & \textbf{0.5261±.0212}  & 0.7891±.0226          & 0.5302±.0765 & 1.0076±.1280          & 0.6471±.0179 & 0.8301±.0328 & 0.7826±.0130 & 4.1335±.0168          & 0.6984±.0118          \\
		& sja   & \textbf{0.0341±.0297}  & 0.1979±.0334          & 0.1965±.0298 & 0.0652±.0001          & 0.0717±.0054 & 0.2031±.0190 & 0.1841±.0004 & 0.0732±.0002          & 0.0768±.0067 \\ \bottomrule         
	\end{tabular}
\label{biao1}
\end{table*}

\begin{table*}[!t]\small\centering\renewcommand{\arraystretch}{1.1}
	\setlength{\tabcolsep}{0.005mm}
	\caption{Predictive performance of each comparing approach (mean $\pm$ std) in terms of Clark distance$\downarrow$, Intersection similarity$\uparrow$ and KL distance$\downarrow$. $\uparrow(\downarrow)$ indicates the larger (smaller) the value, the better the performance. Best results are shown in boldface.}
	\begin{tabular}{ccccccccccc}\toprule
		&       & $I^2LDL$               & In-a          & In-p                  & In-GSC                & LDLLDM                & LDLLC         & LDLSF         & LSRLDL                & SILDL                 \\\toprule
		\multirow{16}{*}{Can.$\downarrow$} & alpha & \textbf{0.6741±.0064}  & 0.6831±.0178  & 0.6857±.0185          & 1.5661±.0046          & 1.0889±.0028          & 1.5497±.1036  & 0.6942±.0006  & 0.7287±.0147          & 0.7314±.0266          \\
		& cold  & \textbf{0.2494±.0126}  & 0.2503±.0004  & 0.2507±.0002          & 0.2544±.0105          & 0.2722±.0025          & 0.3969±.0248  & 0.2540±.0046  & 0.2524±.0038          & 0.2503±.0101          \\
		& dtt   & \textbf{0.1659±.0084}  & 0.1699±.0069  & 0.1703±.0070          & 0.1832±.0123          & 0.2040±.0037          & 0.3906±.0201  & 0.1751±.0029  & 0.1836±.0026          & 0.1756±.0042          \\
		& elu   & \textbf{0.5768±.0056}  & 0.6038±.0108  & 0.5899±.0113          & 1.0834±.0231          & 0.8806±.0034          & 1.1947±.0373  & 0.5982±.0013  & 0.6044±.0039          & 0.6010±.0089          \\
		& heat  & \textbf{0.3552±.0180}  & 0.3816±.0004  & 0.3822±.0004          & 0.4053±.0003          & 0.4439±.0056          & 0.6221±.0033  & 0.3717±.0009  & 0.3704±.0023          & 0.3840±.0081          \\
		& spo5  & \textbf{0.2758±.0010}  & 0.2854±.0048  & 0.2865±.0043          & 0.3174±.0307          & 0.2947±.0017          & 0.3708±.0311  & 0.2832±.0021  & 0.2814±.0002          & 0.2938±.0030          \\
		& spo   & \textbf{0.5011±.0081}  & 0.5312±.0574  & 0.5318±.0582          & 0.5413±.0080          & 0.5636±.0078          & 0.6782±.0251  & 0.5303±.0017  & 0.5155±.0034          & 0.5128±.0026          \\
		& spoem & \textbf{0.1653±.0115}  & 0.1764±.0072  & 0.1767±.0067          & 0.2026±.0080          & 0.1871±.0004          & 0.2195±.0009  & 0.1802±.0046  & 0.1891±.0011          & 0.1805±.0123          \\
		& fli   & \textbf{0.3120±.0004}  & 0.3967±.0172  & 0.7861±.0123          & 0.4322±.0027          & 0.5479±.0041          & 0.8272±.0498  & 0.3308±.0118  & 0.3430±.0021          & 0.3430±.0081          \\
		& twi   & \textbf{0.3533±.0058}  & 0.4253±.0007  & 0.8186±.0158          & 0.5491±.0152          & 0.6414±.0418          & 1.0281±.2072  & 0.3941±.0011  & 0.3772±.0007          & 0.3774±.0030          \\
		& human & \textbf{14.2755±.2803} & 14.4051±.1708 & 14.4169±.0725         & 15.6079±.3993         & 14.7380±.0050         & 17.7585±.9864 & 14.5305±.0994 & 14.4973±.1572         & 14.4151±.0341         \\
		& nat   & \textbf{6.8605±.0740}  & 6.8909±.0034  & 7.1589±.0524          & 6.9039±.0749          & 6.8665±.0046          & 6.9162±.2106  & 6.9895±.0107  & 7.0126±.0078          & 7.0225±.0389          \\
		& ren   & \textbf{7.3045±.0206}  & 7.4292±.0331  & 7.4388±.0054          & 7.4202±.0005          & 7.4242±.0018          & 7.3811±.0266  & 7.4441±.0025  & 7.4673±.0011          & 7.4759±.0095          \\
		& sbu   & \textbf{0.8535±.0550}  & 1.0164±.0207  & 1.0234±.0203          & 0.8873±.0290          & 0.8955±.0077          & 0.8979±.0291  & 0.8889±.0046  & 0.8911±.0031          & 0.9015±.0057          \\
		& scut  & \textbf{2.9152±.0615}  & 3.0020±.0062  & 3.3190±.0249          & 5.8640±3.2863         & 2.9286±.0230          & 2.9871±.0310  & 2.9461±.0509  & 2.9387±.0132          & 2.9672±.0060          \\
		& sja   & \textbf{0.7932±.1258}  & 0.9071±.1877  & 0.9009±.1752          & 0.8641±.0008          & 0.8883±.0318          & 0.9102±.0557  & 0.8553±.0050  & 0.9121±.0045          & 0.8811±.0666          \\\bottomrule 
		\multirow{16}{*}{Cos.$\uparrow$}   & alpha & \textbf{0.9948±.0001}  & 0.9944±.0002  & 0.9944±.0002          & 0.9752±.0002          & 0.9864±.0002          & 0.9764±.0020  & 0.9945±.0000  & 0.9939±.0002          & 0.9937±.0003          \\
		& cold  & 0.9877±.0015           & 0.9879±.0001  & 0.9879±.0002          & 0.9873±.0013          & 0.9855±.0002          & 0.9719±.0022  & 0.9872±.0004  & 0.9876±.0003          & \textbf{0.9882±.0008} \\
		& dtt   & \textbf{0.9943±.0006}  & 0.9942±.0004  & 0.9942±.0004          & 0.9934±.0008          & 0.9914±.0003          & 0.9738±.0035  & 0.9937±.0002  & 0.9929±.0003          & 0.9936±.0004          \\
		& elu   & \textbf{0.9943±.0002}  & 0.9937±.0004  & 0.9940±.0004          & 0.9802±.0008          & 0.9859±.0002          & 0.9769±.0016  & 0.9938±.0000  & 0.9937±.0001          & 0.9938±.0003          \\
		& heat  & 0.9882±.0011           & 0.9869±.0000  & 0.9869±.0001          & 0.9858±.0001          & 0.9824±.0005          & 0.9672±.0021  & 0.9874±.0000  & \textbf{0.9874±.0002} & 0.9864±.0008          \\
		& spo5  & \textbf{0.9740±.0005}  & 0.9740±.0012  & 0.9737±.0010          & 0.9693±.0048          & 0.9717±.0004          & 0.9574±.0058  & 0.9740±.0003  & 0.9743±.0001          & 0.9733±.0003          \\
		& spo   & \textbf{0.9777±.0010}  & 0.9753±.0047  & 0.9752±.0048          & 0.9741±.0001          & 0.9716±.0006          & 0.9599±.0016  & 0.9747±.0002  & 0.9760±.0003          & 0.9756±.0002          \\
		& spoem & \textbf{0.9811±.0021}  & 0.9792±.0023  & 0.9791±.0022          & 0.9750±.0018          & 0.9760±.0001          & 0.9682±.0002  & 0.9780±.0009  & 0.9756±.0003          & 0.9767±.0014          \\
		& fli   & \textbf{0.9960±.0000}  & 0.9909±.0011  & 0.9623±.0006          & 0.9900±.0000          & 0.9836±.0008          & 0.9600±.0045  & 0.9948±.0004  & 0.9952±.0000          & 0.9952±.0002          \\
		& twi   & \textbf{0.9941±.0000}  & 0.9907±.0011  & 0.9463±.0015          & 0.9809±.0010          & 0.9772±.0036          & 0.9444±.0190  & 0.9931±.0001  & 0.9929±.0003          & 0.9929±.0001          \\
		& human & 0.8408±.0021           & 0.8369±.0024  & 0.8338±.0031          & 0.8201±.0018          & \textbf{0.8297±.0000} & 0.7966±.0072  & 0.8336±.0014  & 0.8341±.0033          & 0.8349±.0014          \\
		& nat   & 0.6009±.0056           & 0.6218±.0152  & 0.6226±.0160          & \textbf{0.6242±.0108} & 0.6309±.0017          & 0.5936±.0245  & 0.6013±.0021  & 0.5712±.0022          & 0.5741±.0055          \\
		& ren   & \textbf{0.5595±.0040}  & 0.5772±.0047  & 0.5720±.0008          & 0.5875±.0089          & 0.4646±.0040          & 0.4742±.0145  & 0.4430±.0004  & 0.4185±.0002          & 0.4163±.0026          \\
		& sbu   & 0.9256±.0078           & 0.9201±.0080  & 0.9194±.0084          & 0.9221±.0025          & 0.9188±.0006          & 0.9199±.0049  & 0.9210±.0006  & 0.9188±.0011          & 0.9161±.0029          \\
		& scut  & 0.6677±.0051           & 0.6697±.0082  & 0.6079±.0050          & 0.5998±.0342          & \textbf{0.6963±.0101} & 0.6433±.0194  & 0.6754±.0084  & 0.6639±.0022          & 0.6622±.0070          \\
		& sja   & \textbf{0.9462±.0130}  & 0.9295±.0134  & 0.9306±.0108          & 0.9388±.0002          & 0.9325±.0047          & 0.9259±.0106  & 0.9366±.0005  & 0.9264±.0002          & 0.9262±.0056          \\\bottomrule 
		\multirow{16}{*}{Int.$\uparrow$}   & alpha & \textbf{0.9628±.0003}  & 0.9623±.0011  & 0.9621±.0011          & 0.9152±.0003          & 0.9385±.0002          & 0.9143±.0056  & 0.9617±.0000  & 0.9598±.0007          & 0.9596±.0015          \\
		& cold  & \textbf{0.9388±.0028}  & 0.9385±.0004  & 0.9384±.0003          & 0.9374±.0023          & 0.9325±.0006          & 0.9019±.0050  & 0.9373±.0010  & 0.9376±.0007          & 0.9379±.0025          \\
		& dtt   & \textbf{0.9590±.0019}  & 0.9579±.0015  & 0.9578±.0015          & 0.9546±.0030          & 0.9493±.0008          & 0.9040±.0048  & 0.9567±.0007  & 0.9549±.0006          & 0.9569±.0010          \\
		& elu   & \textbf{0.9594±.0003}  & 0.9574±.0008  & 0.9584±.0008          & 0.9241±.0017          & 0.9363±.0003          & 0.9152±.0026  & 0.9578±.0001  & 0.9574±.0003          & 0.9574±.0008          \\
		& heat  & \textbf{0.9413±.0028}  & 0.9374±.0006  & 0.9373±.0006          & 0.9331±.0002          & 0.9263±.0011          & 0.8969±.0013  & 0.9389±.0001  & 0.9391±.0003          & 0.9368±.0015          \\
		& spo5  & \textbf{0.9108±.0000}  & 0.9073±.0013  & 0.9070±.0011          & 0.8978±.0077          & 0.9046±.0005          & 0.8787±.0112  & 0.9085±.0005  & 0.9089±.0001          & 0.9055±.0003          \\
		& spo   & \textbf{0.9174±.0016}  & 0.9125±.0091  & 0.9124±.0093          & 0.9106±.0006          & 0.9064±.0015          & 0.8874±.0041  & 0.9127±.0003  & 0.9151±.0005          & 0.9153±.0005          \\
		& spoem & \textbf{0.9193±.0054}  & 0.9145±.0031  & 0.9143±.0029          & 0.9031±.0039          & 0.9091±.0003          & 0.8928±.0012  & 0.9126±.0020  & 0.9087±.0003          & 0.9127±.0059          \\
		& fli   & \textbf{0.9609±.0001}  & 0.9480±.0020  & 0.8978±.0010          & 0.9423±.0000          & 0.9298±.0009          & 0.8912±.0069  & 0.9584±.0015  & 0.9570±.0003          & 0.9570±.0010          \\
		& twi   & \textbf{0.9557±.0007}  & 0.9467±.0014  & 0.8839±.0021          & 0.9265±.0011          & 0.9165±.0064          & 0.8662±.0272  & 0.9505±.0001  & 0.9524±.0000          & 0.9523±.0004          \\
		& human & \textbf{0.7880±.0031}  & 0.7843±.0018  & 0.7846±.0001          & 0.7714±.0009          & 0.7798±.0001          & 0.7418±.0138  & 0.7834±.0014  & 0.7837±.0026          & 0.7848±.0000          \\
		& nat   & 0.3857±.0077           & 0.3902±.0042  & \textbf{0.4710±.0034} & 0.3896±.0099          & 0.4020±.0008          & 0.4092±.0180  & 0.3744±.0014  & 0.3602±.0020          & 0.3616±.0054          \\
		& ren   & \textbf{0.3940±.0038}  & 0.3238±.0052  & 0.3361±.0012          & 0.3266±.0059          & 0.2092±.0025          & 0.2478±.0203  & 0.1965±.0005  & 0.1858±.0001          & 0.1840±.0022          \\
		& sbu   & \textbf{0.8481±.0106}  & 0.8329±.0065  & 0.8317±.0068          & 0.8413±.0049          & 0.8399±.0014          & 0.8395±.0057  & 0.8412±.0008  & 0.8405±.0009          & 0.8384±.0006          \\
		& scut  & 0.5136±.0077           & 0.5045±.0059  & 0.4961±.0029          & 0.3734±.0396          & \textbf{0.5271±.0072} & 0.4966±.0104  & 0.5152±.0093  & 0.5071±.0022          & 0.5041±.0034          \\
		& sja   & \textbf{0.8669±.0217}  & 0.8451±.0285  & 0.8463±.0259          & 0.8547±.0007          & 0.8491±.0060          & 0.8441±.0104  & 0.8547±.0004  & 0.8429±.0007          & 0.8463±.0114   \\\bottomrule    
	\end{tabular}
\label{biao2}
\end{table*}

	\begin{table}
	\centering\renewcommand{\arraystretch}{1.2}
	\caption{The distribution distance/similarity measures.}
	\label{PINGJIAZHIBIAO}
	\setlength{\tabcolsep}{8mm}
	\begin{tabular}{@{}ll@{}}
		\toprule
		Measure          & Formula \\ 
		\midrule
		Chebyshev (Che) $\downarrow$       &   $\operatorname{Dis}_1(\boldsymbol{d}, \hat{\boldsymbol{d}})=\max _j\left|d_j-\hat{d}_j\right|$       \\ 
		Clark (Cla) $\downarrow$           &  $\operatorname{Dis}_2(\boldsymbol{d}, \hat{\boldsymbol{d}})=\sqrt{\sum_{j=1}^c \frac{\left(d_j-\hat{d}_j\right)^2}{\left(d_j+\hat{d}_j\right)^2}}$        \\ 
		Canberra (Can) $\downarrow$        &   $\operatorname{Dis}_3(\boldsymbol{d}, \hat{\boldsymbol{d}})=\sum_{j=1}^c \frac{\left|d_j-\hat{d}_j\right|}{d_j+\hat{d}_j}$      \\ 
		Kullback-Leibler (KL)$\downarrow$ &    $\operatorname{Dis}_4(\boldsymbol{d}, \hat{\boldsymbol{d}})=\sum_{j=1}^c d_j \ln \frac{d_j}{\hat{d}_j}$    \\ 
	
		{Cosine} (Cos) $\uparrow$   & $\operatorname{Sim}_1(\boldsymbol{d}, \hat{\boldsymbol{d}})=\frac{\sum_{j=1}^c d_j \hat{d}_j}{\sqrt{\sum_{j=1}^c d_j^2} \sqrt{\sum_{j=1}^c \hat{d}_j^2}}$     \\ 
		Intersection (Inter)$\uparrow$ & $\operatorname{Sim}_2(\boldsymbol{d}, \hat{\boldsymbol{d}})=\sum_i \min \left(\boldsymbol{d}, \hat{\boldsymbol{d}}\right)$ \\ 
		\bottomrule
	\end{tabular}
\end{table}

\begin{table*}[!h]\caption{Summary of the Friedman statistics $F_F$ 
		in terms  of six evaluation metrics, as well as the critical value at a significance level of 0.05 (9 algorithms on 16 datasets). }
	\small
	\centering
	\renewcommand{\arraystretch}{1.1}
	\setlength{\tabcolsep}{0.9mm}
	\begin{tabular}{@{}l|ccccccc@{}}
		\hline
		Critical   Value ($\alpha=0.05$) & Evaluation metric & Chebyshev & Canberra & Cosine & Clark & Intersection & KL \\ \hline
		\multicolumn{1}{c|}{2.460 } & Friedman Statistics $F_F$ & 68.133
		 & 60.983 & 51.484
		  & 61.166
		   & 55.816
		    & 75.066
		     \\ \hline
	\end{tabular}
	\label{F-value}
\end{table*}

\begin{figure*}[!h]
	\centering
	\subfloat[Intersection$\uparrow$]{
		\includegraphics[width=0.3\linewidth]{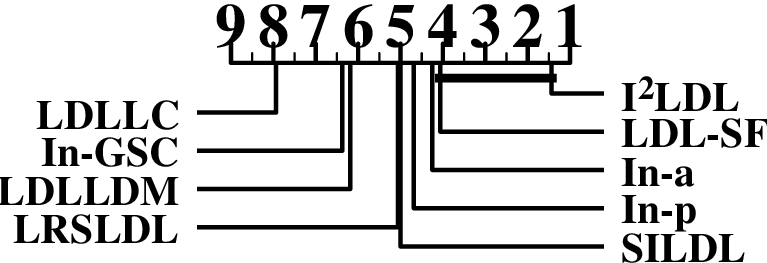}
	}
	\hfil
	\subfloat[Clark $\downarrow$]{
		\includegraphics[width=0.3\linewidth]{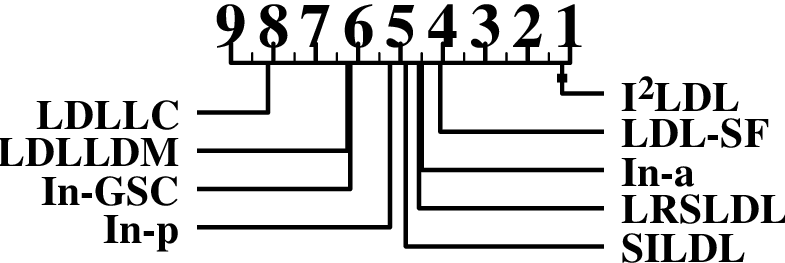}
	}
	\hfil
	\subfloat[Cosine$\uparrow$]{
		\includegraphics[width=0.3\linewidth]{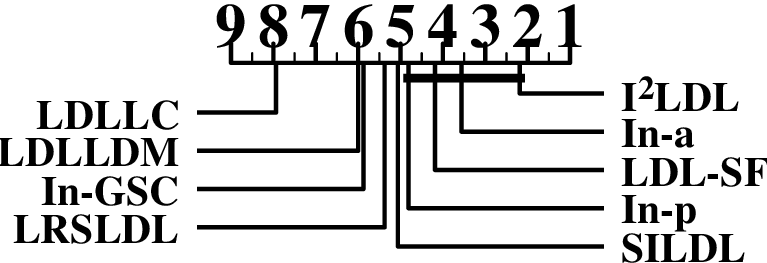}
	}
	\vfil
	\subfloat[Canberra $\downarrow$]{
		\includegraphics[width=0.3\linewidth]{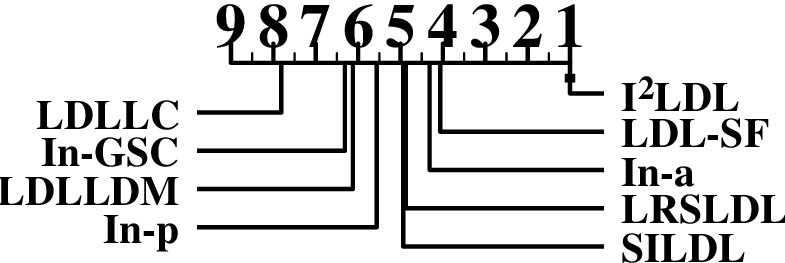}
	}
	\hfil
	\subfloat[Chebyshev$\downarrow$]{
		\includegraphics[width=0.3\linewidth]{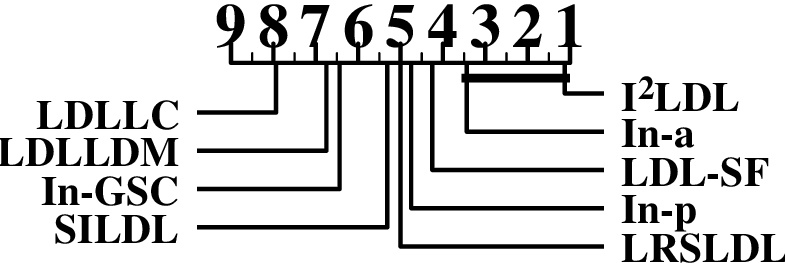}
	}
	\hfil
	\subfloat[KL$\downarrow$]{
		\includegraphics[width=0.3\linewidth]{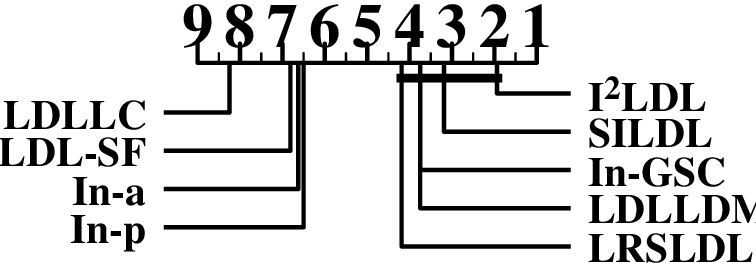}
	}
	
	\caption{CD diagrams of the comparing algorithms in terms of each evaluation criterion. For the tests, CD equals 2.25 at 0.05 significance level.}
	\label{CD1}
\end{figure*}

\begin{figure*}[!htb]
	\centering
	\subfloat[Chebyshev]{\includegraphics[width=0.3\textwidth]{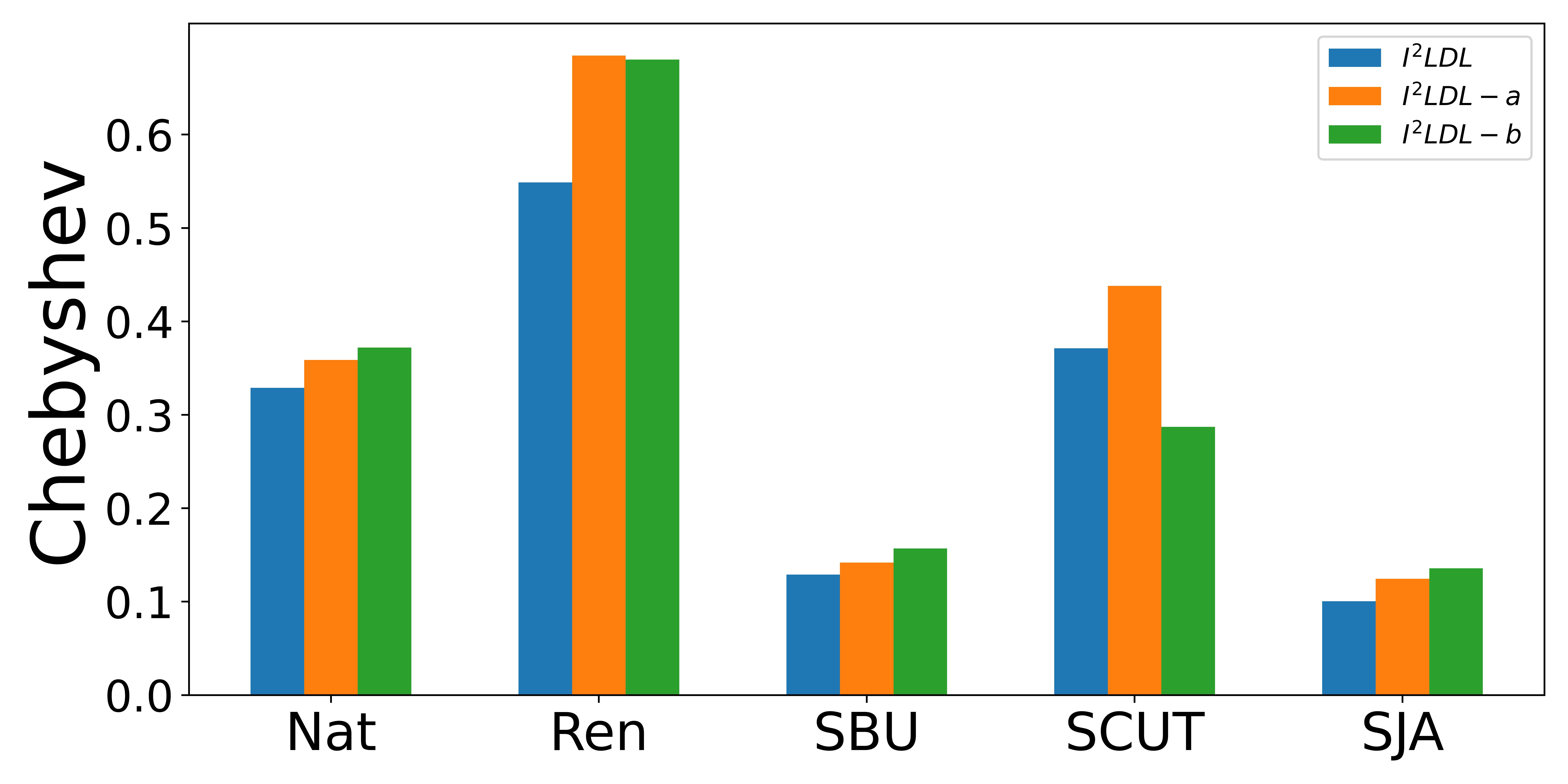}}
	\hfill
	\subfloat[Clark]{\includegraphics[width=0.3\textwidth]{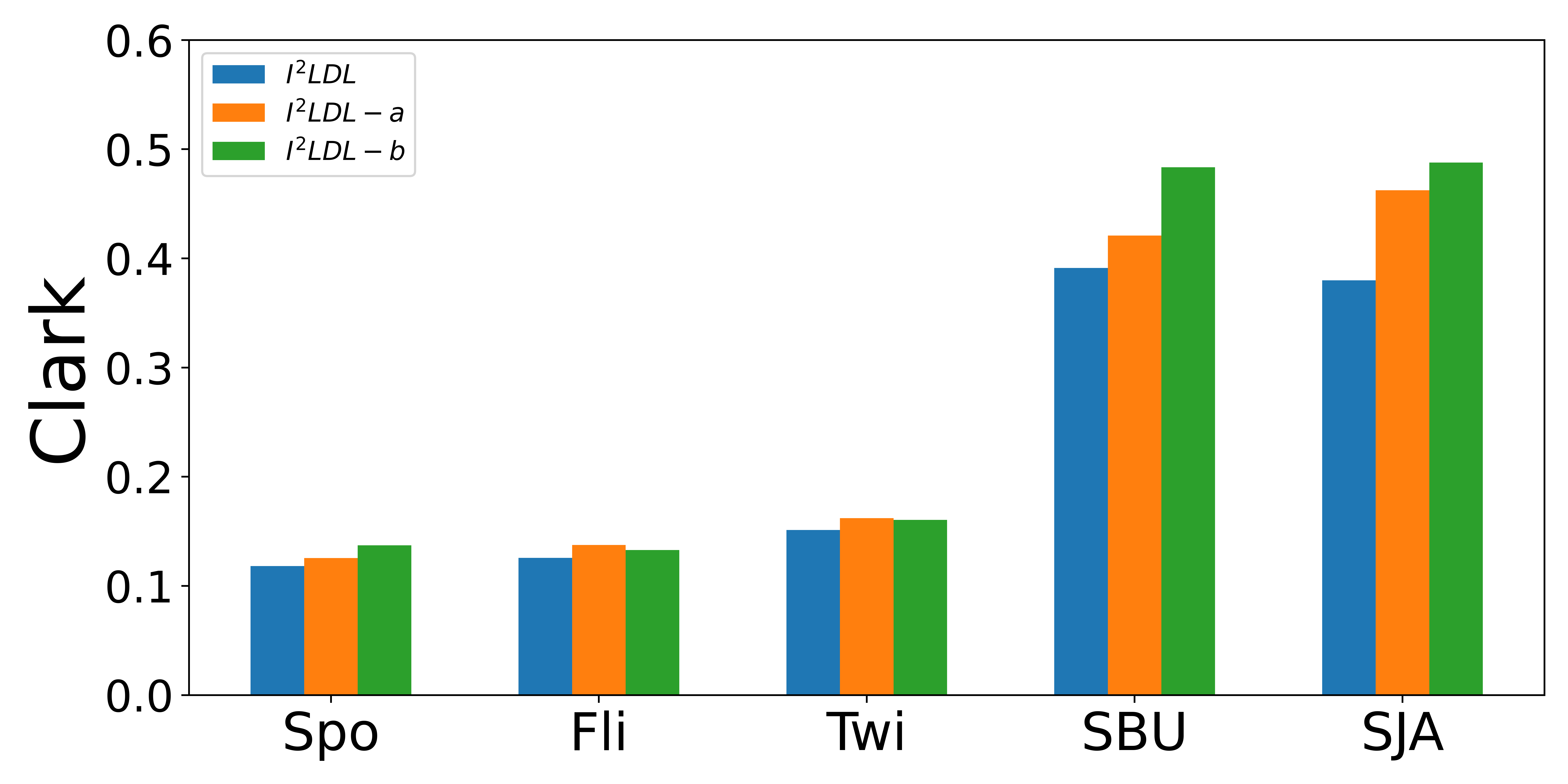}}
	\hfill
	\subfloat[Canberra]{\includegraphics[width=0.3\textwidth]{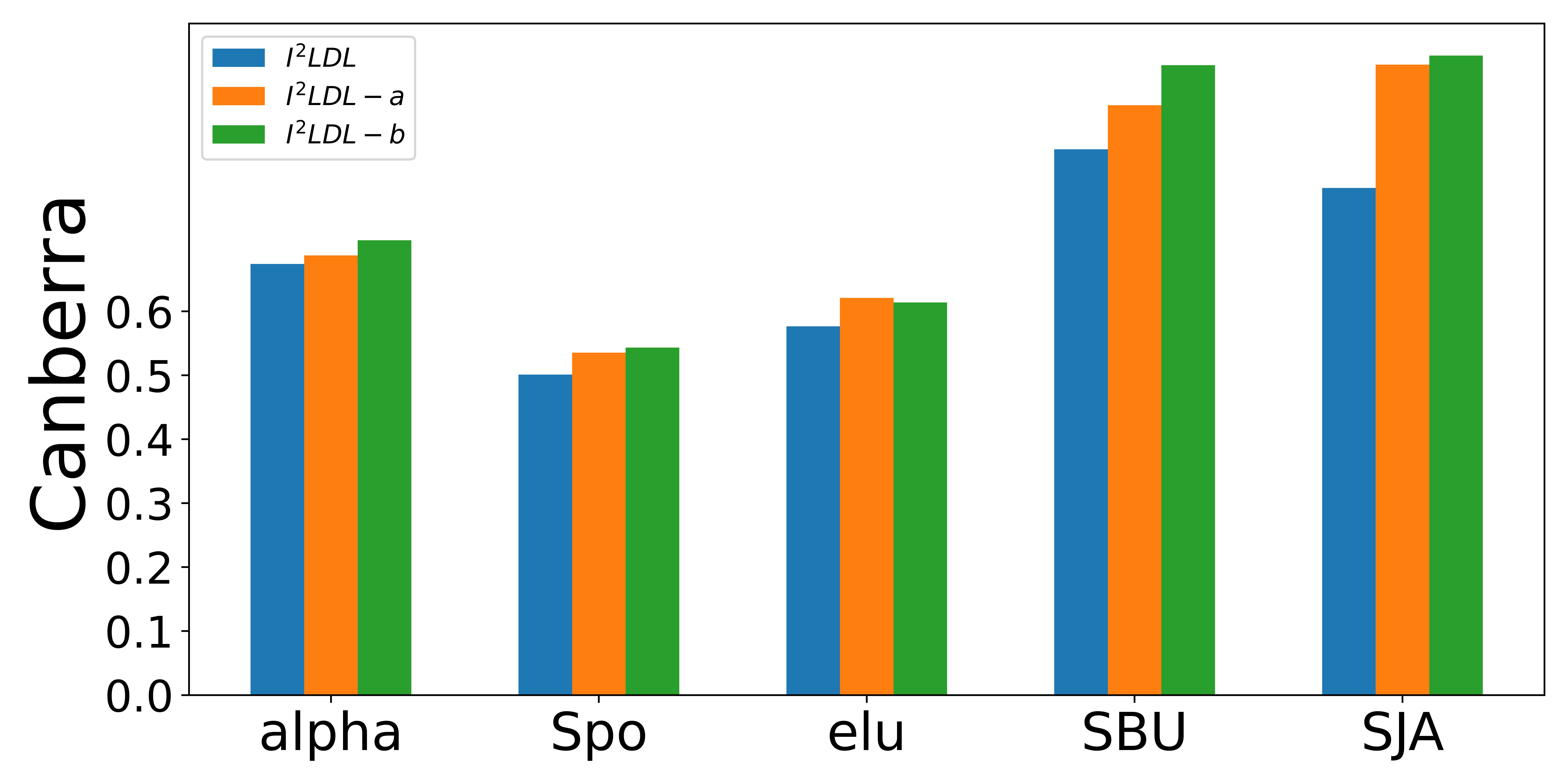}}
	
	\vfil
	\subfloat[Cosine]{\includegraphics[width=0.3\textwidth]{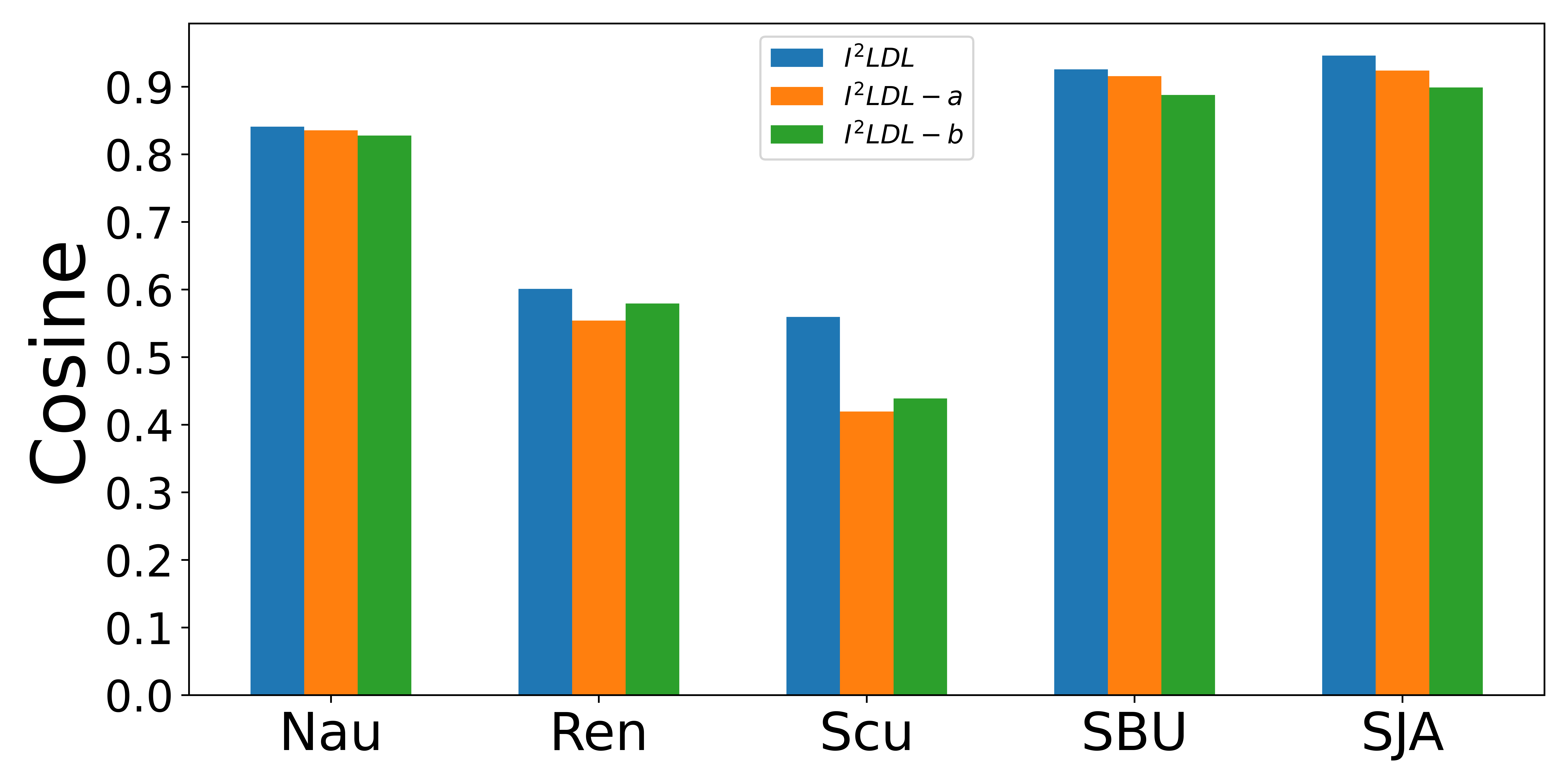}}
	\hfill
	\subfloat[Intersection]{\includegraphics[width=0.3\textwidth]{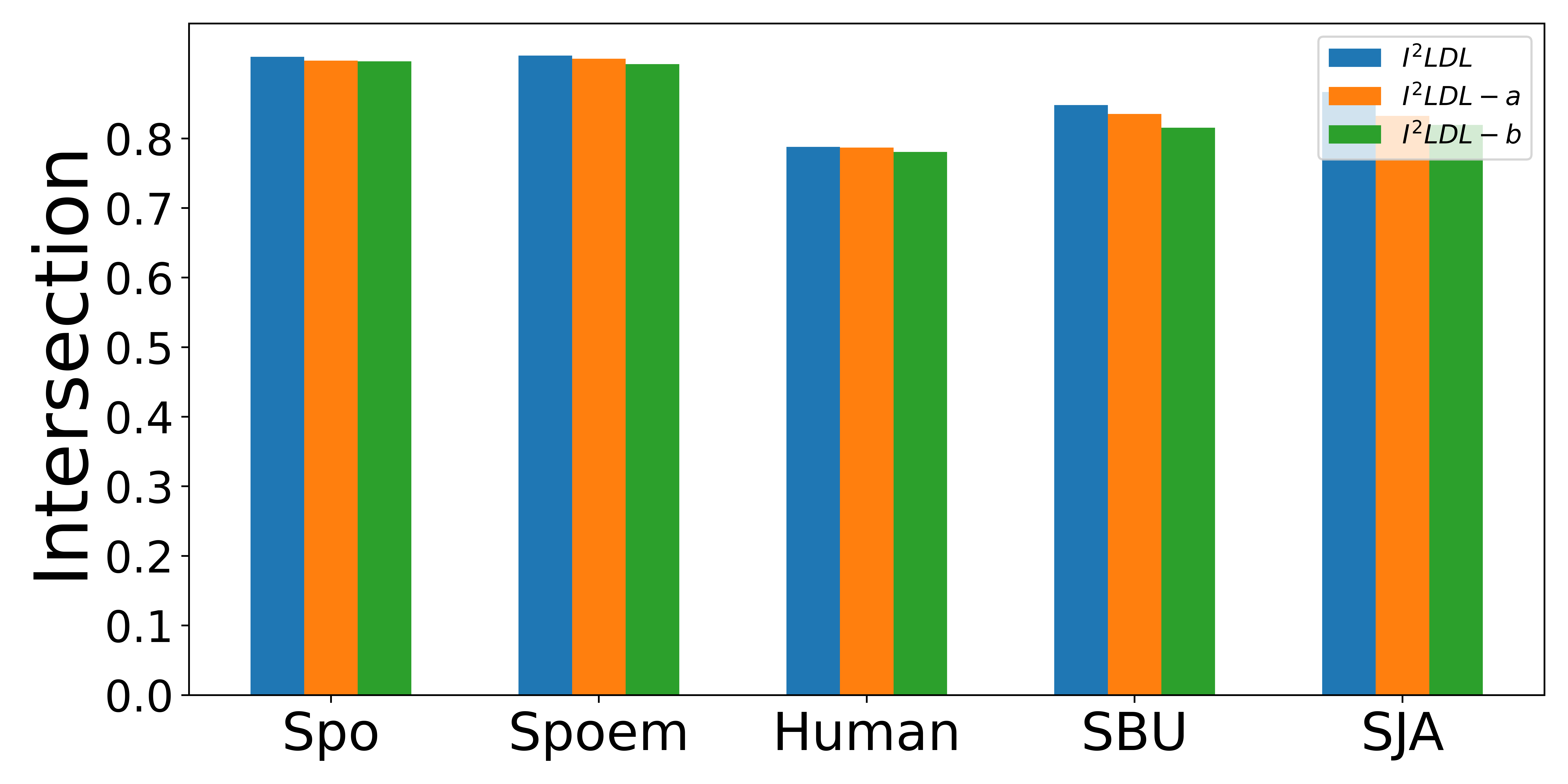}}
	\hfill
	\subfloat[KL]{\includegraphics[width=0.3\textwidth]{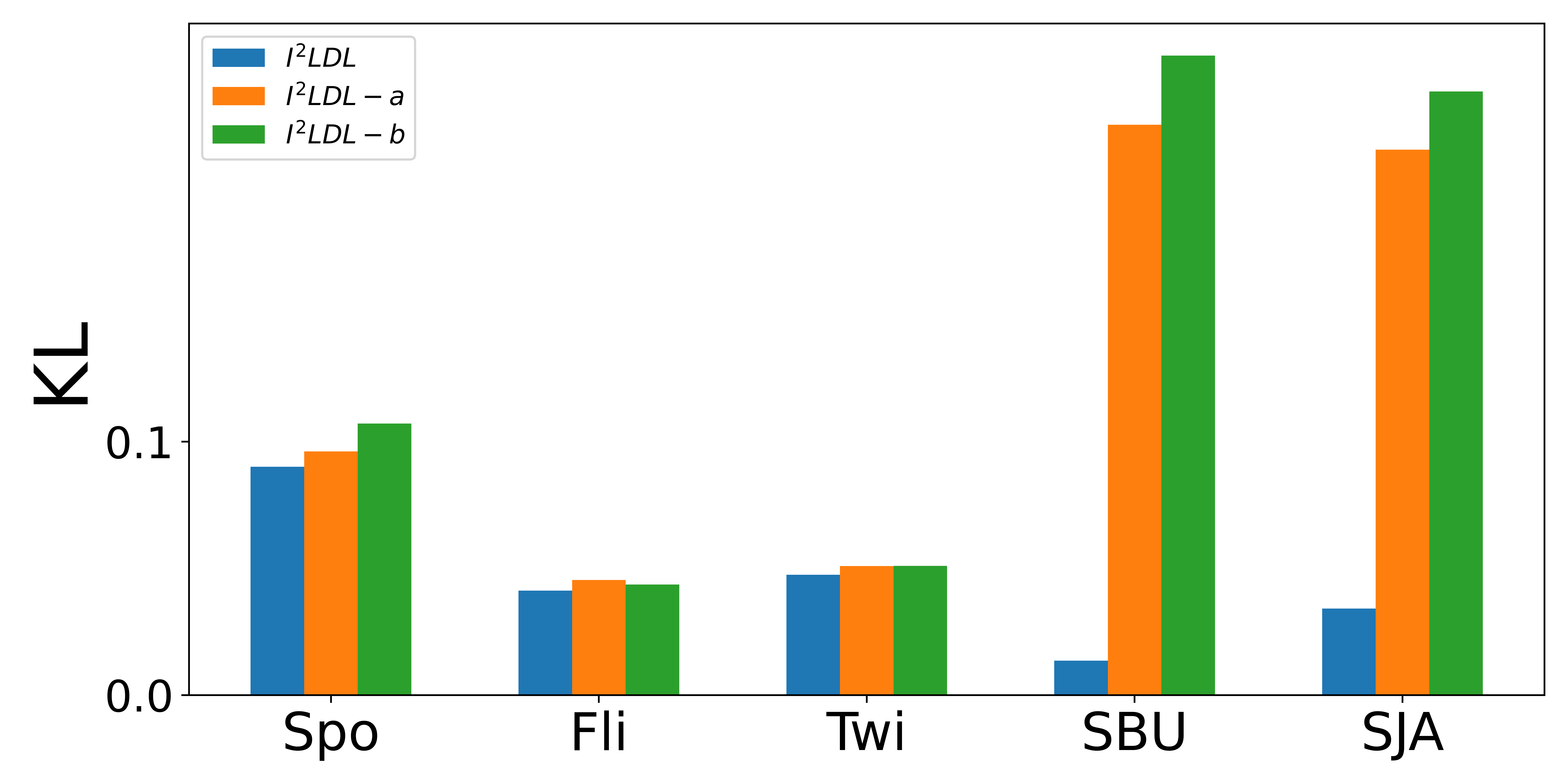}}
	
	\caption{Comparison of different metrics across datasets for $I^2LDL$, $I^2LDL-a$, and $I^2LDL-b$ models.}
	\label{fig:comparison}
\end{figure*}

	\begin{figure*}[!h]
		\centering
		\subfloat[]{
			\includegraphics[width=0.3\linewidth]{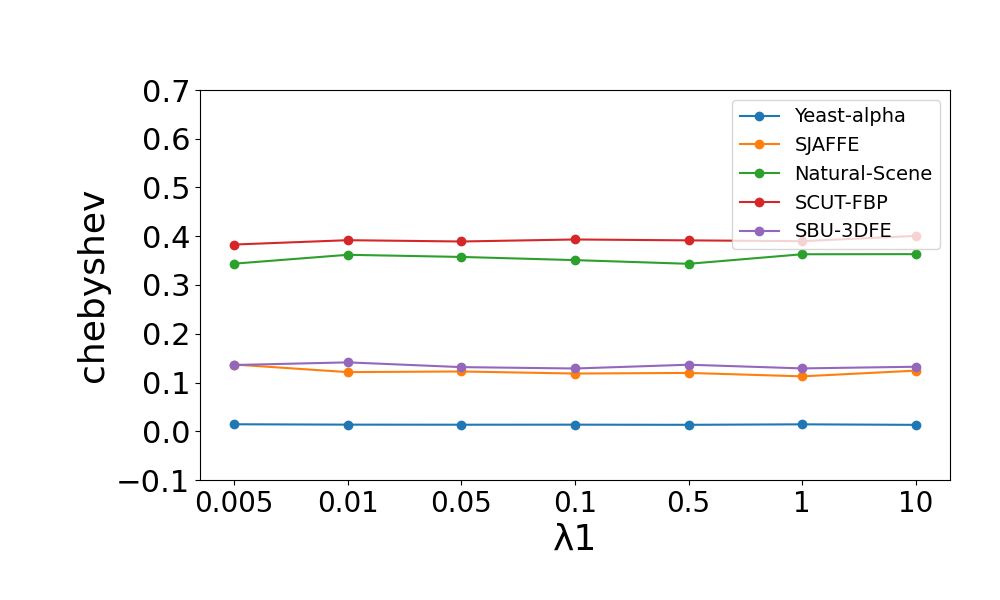}
		}
		\hfill
		\subfloat[]{
			\includegraphics[width=0.3\linewidth]{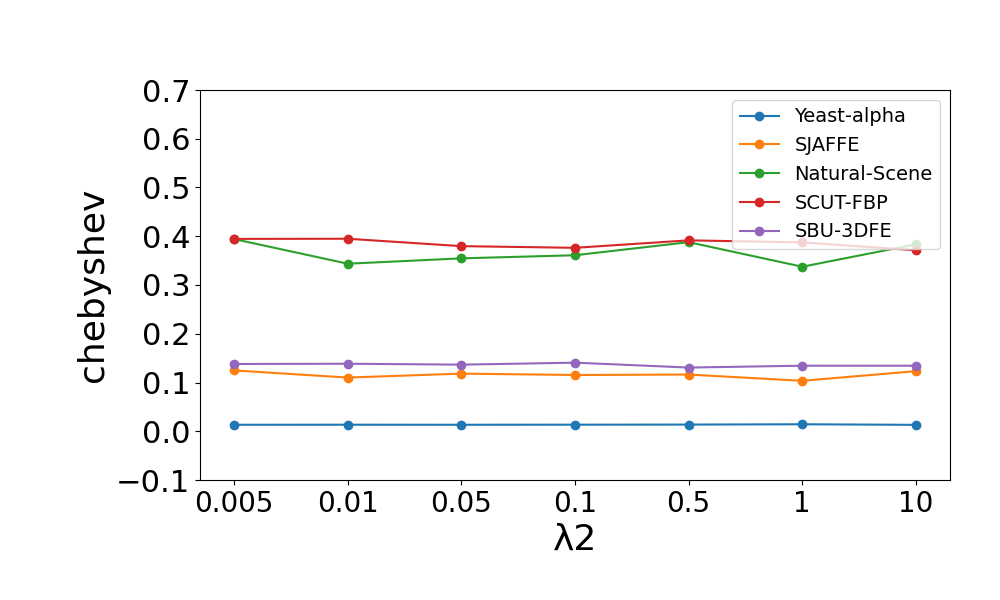}
		}
		\hfill
		\subfloat[]{
			\includegraphics[width=0.3\linewidth]{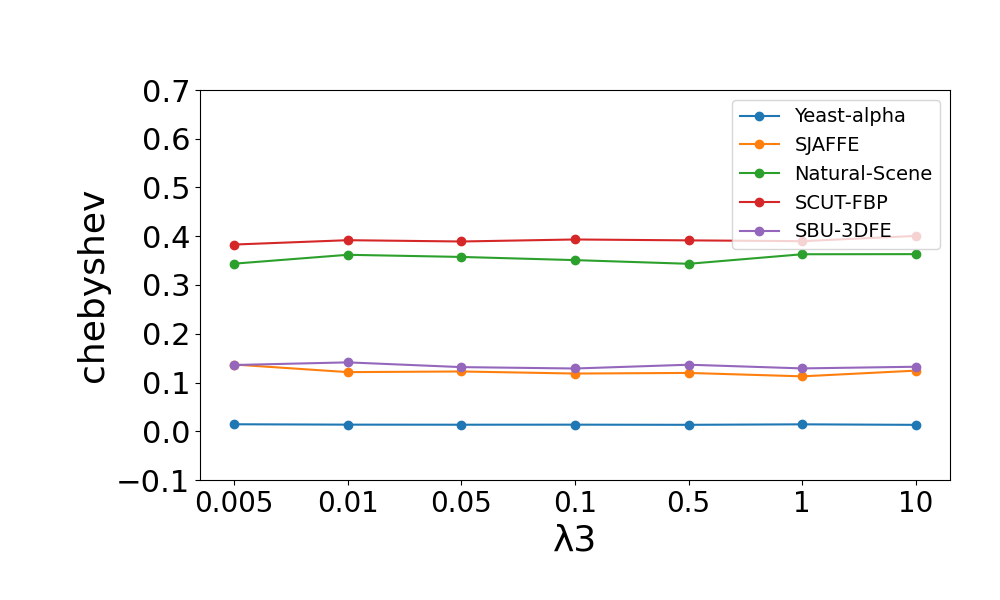}
		}
		
		\vfil
		\subfloat[]{
			\includegraphics[width=0.3\linewidth]{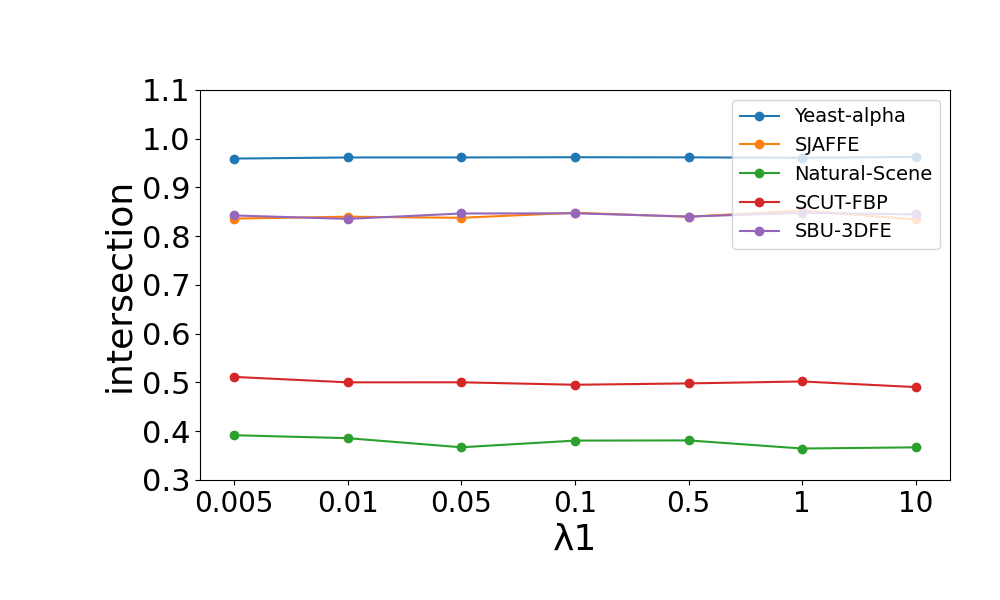}
		}
		\hfill
		\subfloat[]{
			\includegraphics[width=0.3\linewidth]{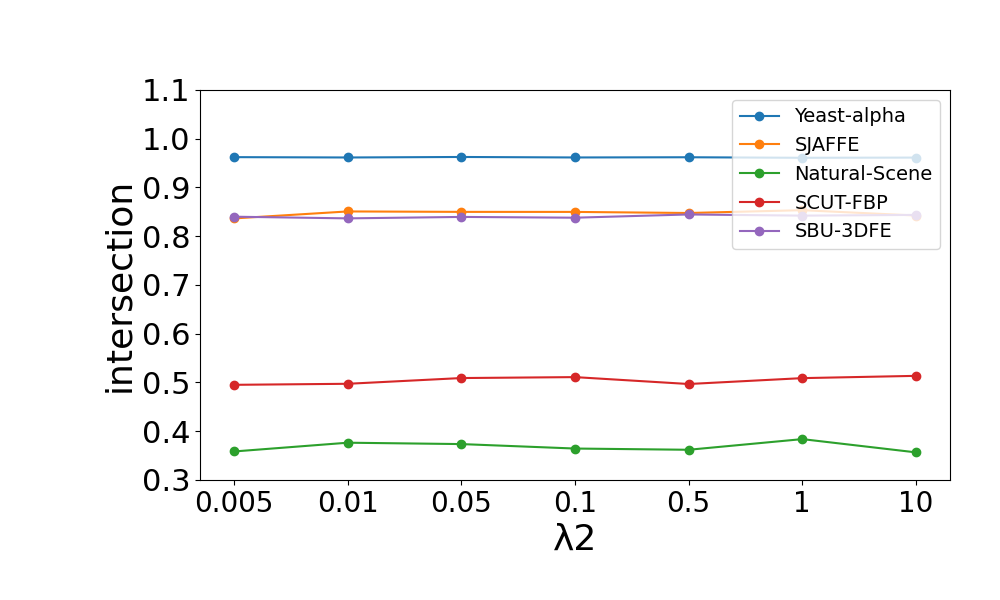}
		}
		\hfill
		\subfloat[]{
			\includegraphics[width=0.3\linewidth]{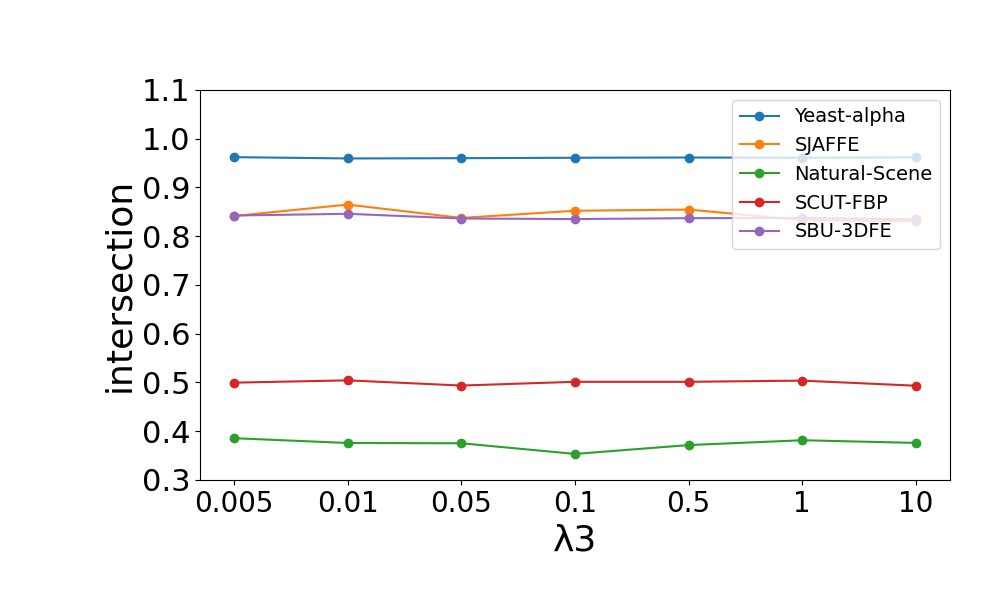}
		}
		
		\caption{Convergence Curves for Different $\lambda$ Values}
		\label{fig:convergence}
	\end{figure*}

\begin{figure}[!h]
	\centering
	\subfloat[Natural Scene]{\includegraphics[width=0.23\textwidth]{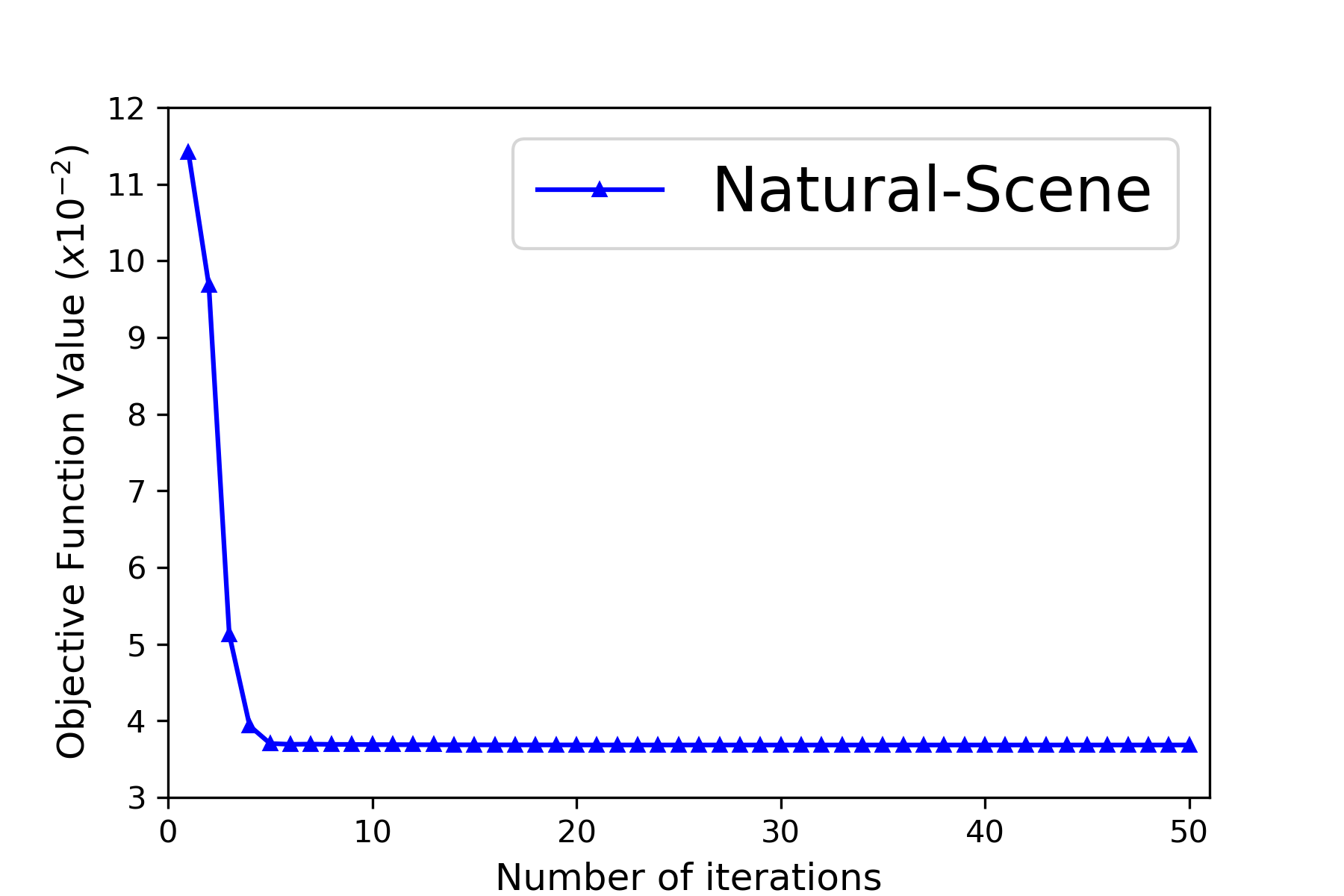}}
	\hspace{0.01\textwidth} 
	\subfloat[SCUT-FBP]{\includegraphics[width=0.23\textwidth]{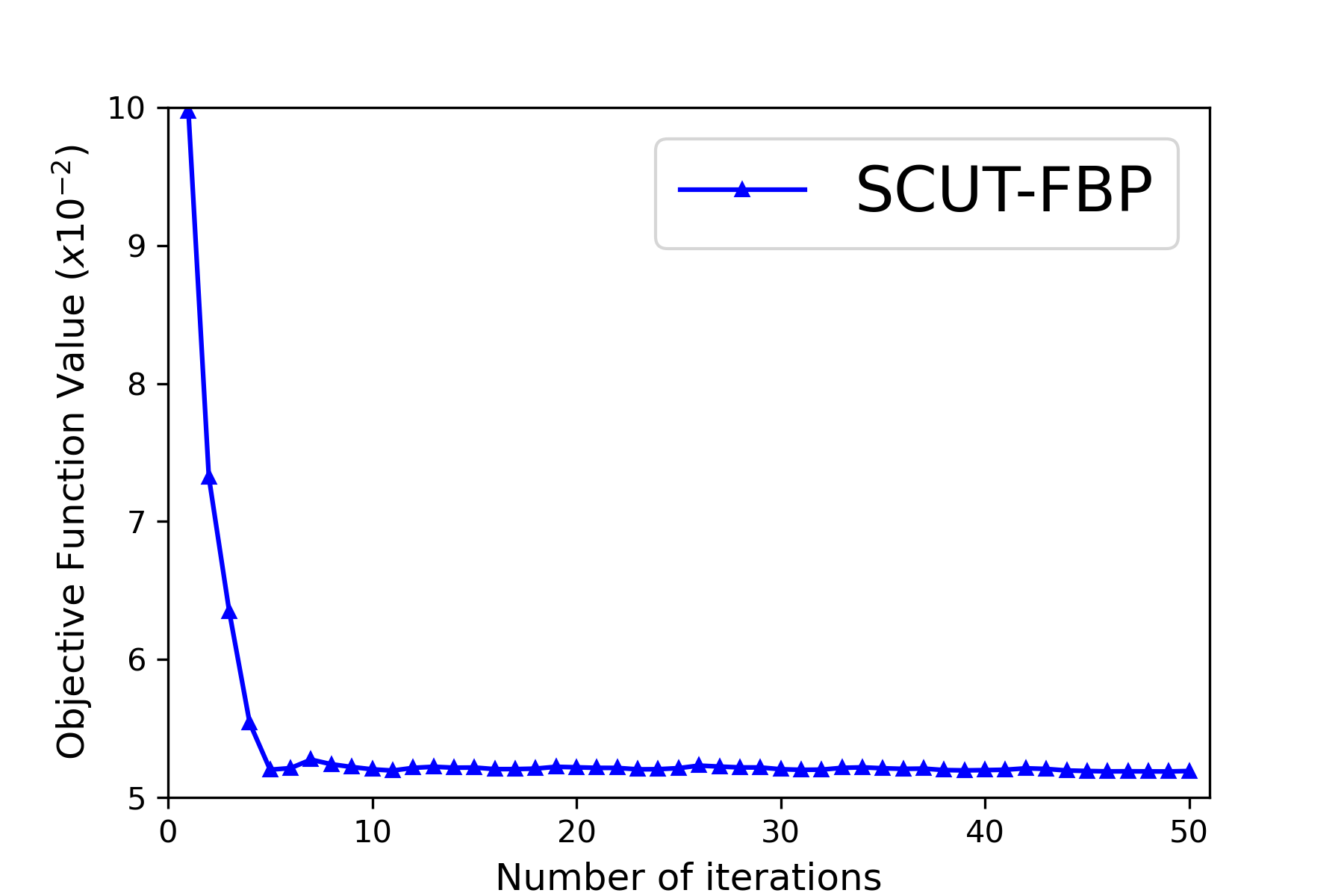}}
	\hspace{0.01\textwidth}
	\subfloat[SJAFFE]{\includegraphics[width=0.23\textwidth]{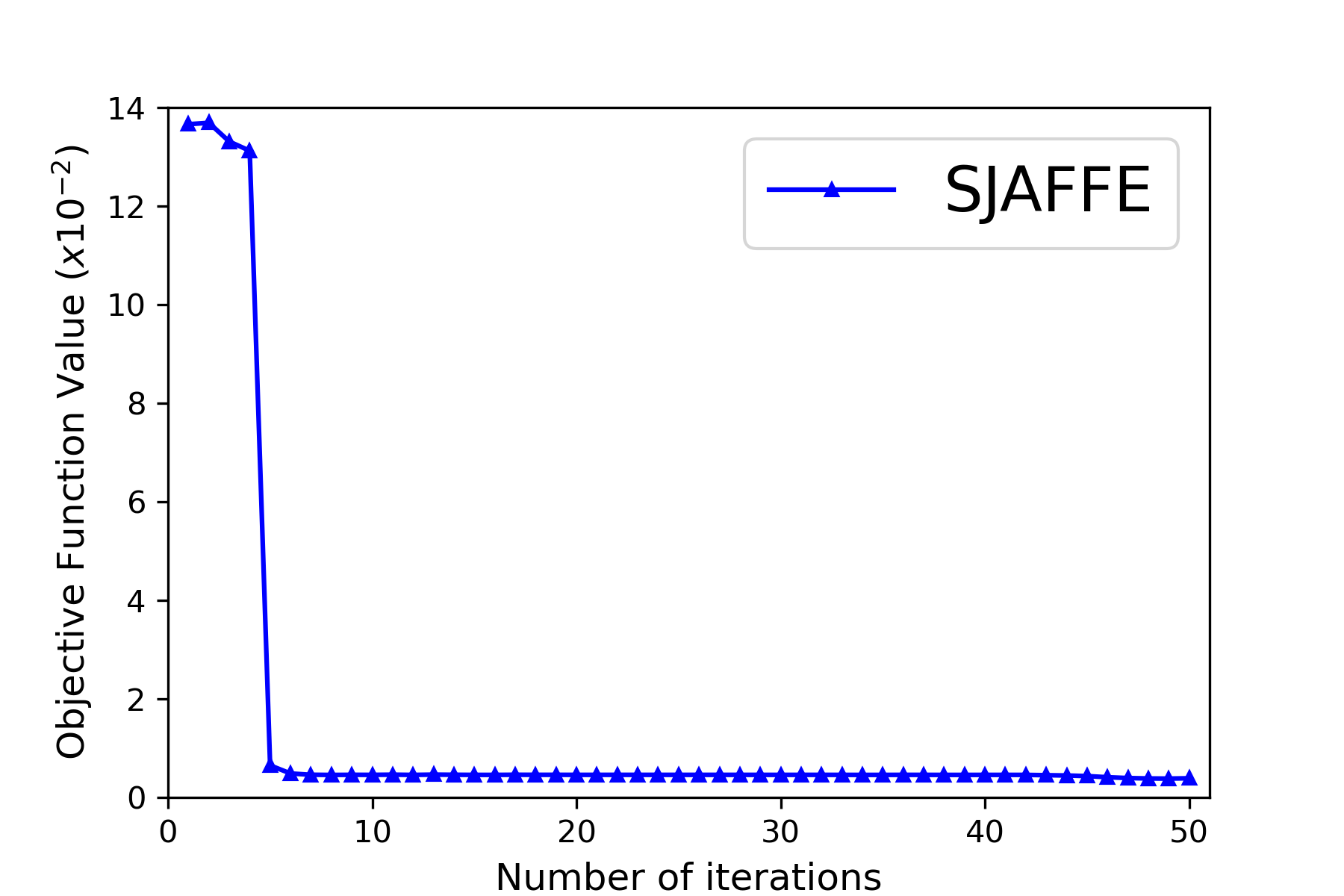}}
	\hspace{0.01\textwidth}
	\subfloat[Twitter]{\includegraphics[width=0.23\textwidth]{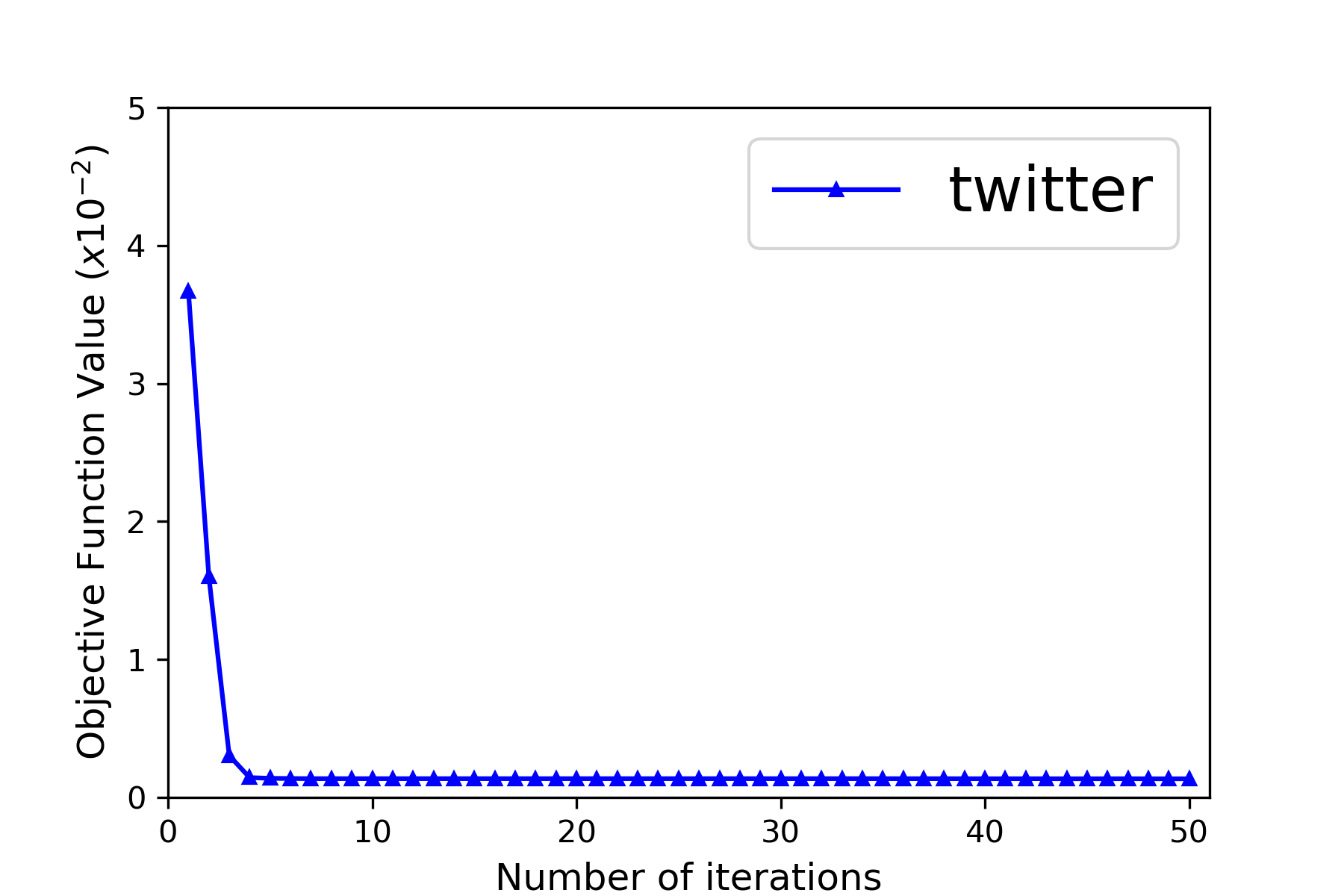}}
	\caption{Convergence Curves for Different Datasets}
	\label{convergence_curves}
\end{figure}

\begin{table}[!h]
	\centering
	\caption{Comparison of $I^2LDL$ with $I^2LDL-a$ and $I^2LDL-b$ across different metrics}
	\renewcommand{\arraystretch}{1.5}  
	\setlength{\tabcolsep}{5pt}  
	\begin{tabular}{l c c}
		\toprule
		\textbf{Metric} & \textbf{$I^2LDL$ vs $I^2LDL-a$} & \textbf{$I^2LDL$ vs $I^2LDL-b$} \\
		\midrule
		Chebyshev   & \textbf{win [5.31e-04]}  & \textbf{win [4.38e-04]}  \\
		Clark       & \textbf{win [5.31e-04]}  & \textbf{win [4.38e-04]}  \\
		Canberra    & \textbf{win [1.92e-03]}  & \textbf{win [4.38e-04]}  \\
		KL          & \textbf{win [4.38e-04]}  & \textbf{win [4.38e-04]}  \\
		Cosine      & \textbf{win [1.53e-03]}  & \textbf{win [6.10e-05]}  \\
		Intersection & \textbf{win [4.38e-04]}  & \textbf{win [4.38e-04]}  \\
		\bottomrule
	\end{tabular}
	\label{tab:comparison}
\end{table}

\subsubsection{Experimental Settings}

In our experiments, we simulate both label imbalance and incomplete label distribution data, following the setup from \cite{zhao2023imbalanced}. Specifically, the training set is designed to exhibit significant label imbalance, where the imbalance factor $\gamma$, defined as the ratio between the most frequent and least frequent labels, is greater than 10. On this imbalanced training set, we further introduced missing labels by randomly setting 50\% of the label distribution matrix elements to be missing, with a fixed missing rate $\omega = 50\%$. The test set, on the other hand, remains balanced and fully observed to evaluate the performance of the model on more uniform label distributions. The training and test sets are split as 90\% and 10\% of the total data, respectively. After training, we assess the difference between the ground-truth label distribution values in the test set and the predicted values. All experiments are conducted with 10-fold cross-validation, and we report the mean and variance of the results.

\subsubsection{Evaluation Metrics}
We use a combination of six metrics to evaluate the performance of the LDL algorithms. These metrics include 5 distance-based measures, Chebyshev, Clark, Kullback-Leibler (KL), and  Canberra,  and two similarity-based measure Cosine and Intersection. 
The formulas are presented in Table II, where $\boldsymbol{d}$ represents the ground-truth label distribution and $\hat{\boldsymbol{d}}$ is the predicted label distribution. For distance-based measures, lower values indicate better performance ($\downarrow$), while for similarity-based measures, higher values indicate better performance ($\uparrow$).

\subsection{Comparative Studies}

\subsubsection{Comparison with LDL algorithms}
We compare  $I^2LDL$  with eight other LDL algorithms. Among them, five are incomplete LDL methods (i.e., In-a \cite{xu2017incomplete}, In-p \cite{xu2017incomplete}, In-GSC \cite{teng2021incomplete}, LDLLDM \cite{wang2021label_2}, and SIDL \cite{zhang2022safe}), two are standard LDL methods (i.e., LDLLC \cite{jia2018label} and LDLSF \cite{ren2019label}), and one is an inaccurate LDL method (LSRLDL \cite{kou}).  For $I^2LDL$, the trade-off parameters $\alpha$, $\beta$, and $\gamma$ were fine-tuned within the set $\{0.005, 0.01, 0.05, 0.1, 0.5, 1, 10\}$. For the incomplete LDL methods (In-a, In-p, In-GSC), the parameter settings followed those described in their respective original papers. Regarding LDLSF and LDLLC, the models were trained on incomplete label distribution datasets and then evaluated on real-world datasets. Their parameters were also set as per the configurations detailed in the respective papers. The competing methods are summarized as follows:

\begin{itemize}
	
	\item \textbf{LSR-LDL} \cite{kou}: Improves noise management by addressing inaccuracies specific to individual instances within label distributions.
	
	\item \textbf{LDLLC} \cite{jia2018label}: Uses the Pearson correlation to model label correlation in the learning process. 
	
	\item \textbf{LDLSF} \cite{ren2019label}: Proposes to learn label-specific features for each label to improve the performance of learning label distribution. 
	
	\item \textbf{In-a and In-p} \cite{xu2017incomplete}: Utilizes trace-norm regularization and alternating direction methods, effectively leveraging low-rank label correlations.
	
	\item \textbf{LDLLDM} \cite{wang2021label_2}: Exploits both global and local label correlation by learning the label distribution manifold, capable of handling the incomplete LDL problem.
	
	\item \textbf{In-GSC} \cite{teng2021incomplete}: Improves Incomplete LDL by utilizing global sample correlations, capturing the relationships between samples to mitigate the loss of incomplete label information.
	
	\item \textbf{SIDL} \cite{zhang2022safe}: A robust LDL algorithm designed for scenarios with inaccurate labels. It enhances predictive performance by incorporating a safe learning mechanism that minimizes the impact of noise and uncertainty in the label distributions.
	
\end{itemize}

Tables \ref{biao1}  and \ref{biao2}  report the experimental results of the comparing algorithms in terms of six evaluation metrics, where each dataset is denoted by its first three letters. First, the Friedman test \cite{demvsar2006sta} is employed to statistically compare whether there are performance differences among the comparing algorithms. Table \ref{F-value} summarizes the Friedman statistics $F_F$ for each evaluation metric and the corresponding critical value at significance level of 0.05.
From Table \ref{F-value}, the hypothesis that all algorithms perform the same is rejected. 
 
Next, the Bonferroni-Dunn test \cite{dunn1961multiple} is conducted to further investigate the relative performance of  $I^2LDL$ against other approaches. For the tests, the critical difference (CD) \cite{dunn1961multiple} equals 2.25. Fig. \ref{CD1} presents the CD diagrams \cite{demvsar2006sta} for each evaluation metric. In each sub-figure, the average rank of each comparing algorithm is marked along the axis with lower ranks to the right; a thick line connects  $I^2LDL$ and any comparing algorithm if the difference between their average ranks is less than one CD.
According to the above results, we can make the following observations:
\begin{itemize}
	\item \textbf{Performance on Different Metrics:} 
	In terms of \textit{Intersection} and \textit{Cosine} metrics, \textbf{I$^2$LDL} significantly outperforms other compared algorithms, including both incomplete LDL methods and traditional LDL approaches. The performance gap is especially noticeable under the \textit{Intersection} metric, where \textbf{I$^2$LDL} consistently achieves better rankings. For the \textit{Canberra} metric, \textbf{I$^2$LDL} also exhibits stable performance and ranks highly among all compared methods, highlighting its robustness.
	
	\item \textbf{Stability Across Datasets:} 
	The ranking stability of \textbf{I$^2$LDL} across different metrics indicates the robustness of the proposed method. Its strong performance across datasets with varying degrees of imbalance and missing labels demonstrates the generalization ability of the model. The consistent top rankings suggest that \textbf{I$^2$LDL} can effectively handle both incomplete and imbalanced label distributions.
	
	\item \textbf{Comparison with Existing Methods:} 
	Compared with methods like \textbf{LDLLC} and \textbf{LDLLDM}, \textbf{I$^2$LDL} achieves better rankings overall, especially in scenarios with high label imbalance and incomplete data, showcasing the superiority of \textbf{I$^2$LDL} in addressing real-world label distribution challenges.

\end{itemize}

\subsection{Further Analysis}
\subsubsection{Ablation Study}

This section examines the impact of the low-rank and sparsity constraints in the proposed \( I^2LDL \) model. To do this, we compare the full \( I^2LDL \) model with two simplified versions: \( I^2LDL-a \), which omits the low-rank constraint, and \( I^2LDL-b \), which omits the sparsity constraint.

\begin{itemize}
	\item \textbf{Effect of Low-Rank Constraint:} Removing the low-rank constraint in \( I^2LDL-a \) results in a significant performance drop, as shown by the decrease in metrics such as Clark and Cosine (Table VI). This suggests that the low-rank constraint is crucial for capturing the underlying structure of the frequent (head) labels. It helps reduce noise and overfitting by focusing on the most relevant patterns in the data. Without this constraint, the model struggles to generalize, particularly on datasets with complex label distributions.
	
	\item \textbf{Effect of Sparsity Constraint:} The sparsity constraint plays a key role in managing the underrepresented (tail) labels. The removal of this constraint in \( I^2LDL-b \) leads to a performance decrease, but the effect is less severe than in \( I^2LDL-a \). The cardinality constraint ensures that the model remains sensitive to sparse labels, preventing it from overfitting on the more frequent labels and ensuring better generalization to rare labels.
\end{itemize}

In conclusion, both the low-rank and sparsity constraints are essential for achieving optimal performance. The low-rank constraint is particularly important for modeling the structure of frequent labels, while the sparsity constraint improves the handling of less frequent labels.

\subsubsection{Parameter Sensitivity Analysis}

We analyze the sensitivity of \textit{I\(^2\)LDL} to key trade-off parameters \(\lambda_1\), \(\lambda_2\), and \(\lambda_3\). Fig. \ref{fig:convergence} shows the performance variations for the Chebyshev distance and Intersection similarity metrics.

\begin{itemize}
	\item \textbf{Impact of \(\lambda_1\)}: Fig. 2(a) and 2(d) show that model performance remains stable across most \(\lambda_1\) values (0.005-1), with minor fluctuations in Yeast-alpha and SCUT-FBP, indicating \(\lambda_1\) has limited impact on the overall model behavior.
	
	\item \textbf{Impact of \(\lambda_2\)}: As seen in Fig. 2(b) and 2(e), the model performs consistently across datasets, with slight variations in SCUT-FBP and SBU-3DFE using Chebyshev distance. Overall, \(\lambda_2\) does not significantly affect performance.
	
	\item \textbf{Impact of \(\lambda_3\)}: Fig. 2(c) and 2(f) reveal that \(\lambda_3\) has minimal impact across datasets, especially in terms of the Intersection similarity metric, indicating that the model is insensitive to \(\lambda_3\).
\end{itemize}

In conclusion, \textit{I\(^2\)LDL} is robust across a wide range of parameter values, simplifying hyperparameter tuning and ensuring practical applicability.

\subsubsection{Convergence Analysis}

We present the convergence analysis of \textit{I\(^2\)LDL} in Fig. \ref{convergence_curves}, showing the objective function value across iterations for Natural Scene, SCUT-FBP, SJAFFE, and Twitter datasets.

\begin{itemize}
	\item \textbf{Natural Scene:} In Fig. 3(a), the objective function value decreases sharply in the first 10 iterations, indicating rapid convergence.
	\item \textbf{SCUT-FBP:} Fig. 3(b) shows a similar pattern, with convergence achieved after approximately 10 iterations, confirming the efficiency of the method.
	\item \textbf{SJAFFE:} As seen in Fig. 3(c), the method converges quickly despite a higher initial value, stabilizing after 10 iterations.
	\item \textbf{Twitter:} Fig. 3(d) also demonstrates fast convergence within the first few iterations, aligning with the other datasets.
\end{itemize}

Overall, the convergence curves confirm that \textit{I\(^2\)LDL} exhibits fast, stable convergence across diverse datasets, making it efficient and suitable for large-scale applications.

\section{Conclusion}

In this paper, we introduced \textbf{I\(^2\)LDL}, a new framework designed to address the dual challenges of incomplete and imbalanced label distribution learning. By leveraging a low-rank and sparse decomposition of the label distribution matrix, our method effectively models both frequent and rare labels. Extensive experiments on 16 real-world datasets showed that \textbf{I\(^2\)LDL} outperforms existing LDL and Incomplete LDL methods, demonstrating strong predictive accuracy and robustness. Our analyses further confirmed the stability of our approach.

\bibliographystyle{IEEEtran}
\bibliography{IEEEabrv,Bibliography}

\section{Appendix}

In this appendix, we summarize the definitions of \( \mathbf{H}_m \), \( \mathbf{f}_m \), \( \mathbf{H}_z \), and \( \mathbf{f}_z \) used in the subproblems of the optimization process.

\subsection{Definition of \( \mathbf{H}_m \) and \( \mathbf{f}_m \)}

The matrix \( \mathbf{H}_m \) and vector \( \mathbf{f}_m \) are defined as follows:
\begin{equation}
	\mathbf{H}_m = \text{diag} \left( \lambda_1 + \mu \right) \mathbf{\Omega}_i + \rho_2 \overline{\mathbf{\Omega}}_i + 2\lambda_2 (\mathbf{X}^1)^\top (\mathbf{X}^1)
\end{equation}
\begin{equation}
	\mathbf{f}_m = \left( \mathbf{\Lambda}_2^k \right)_i - \left( \mathbf{D}_i \odot \mathbf{\Omega}_i - \mathbf{Z}_i^{k+1} \right) \odot \mathbf{\Omega}_i + \left( \mathbf{\Lambda}_2^k - \mu \mathbf{G}_i^k \right) \odot \overline{\mathbf{\Omega}}_i
\end{equation}

Here, \( \mathbf{H}_m \) is associated with the low-rank structure of the normal label distribution, and \( \mathbf{f}_m \) includes Lagrange multipliers and penalty terms used to update the sparse matrix.

\subsection{Definition of \( \mathbf{H}_z \) and \( \mathbf{f}_z \)}

Similarly, the matrix \( \mathbf{H}_z \) and vector \( \mathbf{f}_z \) are defined as:
\begin{equation}
	\mathbf{H}_z = \text{diag} \left( \mathbf{\Lambda}_1^k + \rho_1 \right) \mathbf{\Omega}_i + \rho_1 \overline{\mathbf{\Omega}}_i
\end{equation}
\begin{equation}
	\mathbf{f}_z = \left( \mathbf{\Lambda}_1^k \right)_i - \left( \mathbf{D}_i \odot \mathbf{\Omega}_i - \left( \mathbf{X} \mathbf{H}^{k+1} \right)_i \right) \odot \mathbf{\Omega}_i - \rho_1 \left( \mathbf{X} \mathbf{U}^{k+1} \mathbf{V}^{k+1} \right)_i
\end{equation}

In this case, \( \mathbf{H}_z \) and \( \mathbf{f}_z \) are related to the low-rank approximation and help ensure that the sparse components of the label distribution matrix are effectively updated.

\begin{IEEEbiography}[{\includegraphics[width=1in,height=1.35in,clip,keepaspectratio]{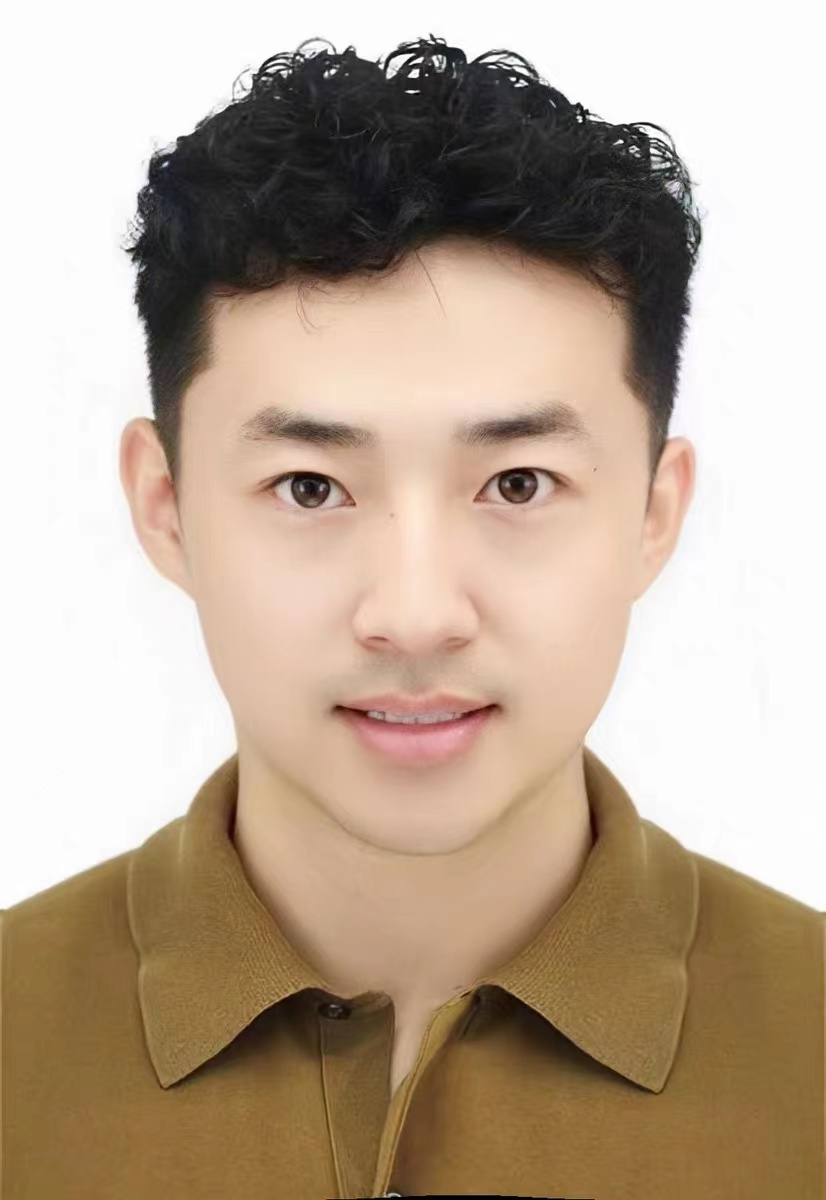}}]{Zhiqiang Kou} received the M.Sc. degree in computer science from Xinjiang University , Urumqi, China, in 2021. He is current a Ph.D. student in the School of Computer Science and Engineering at Southeast University, Nanjing, China. 
	
His research interests include multi-label learning and weakly  supervised learning.
\end{IEEEbiography}

\begin{IEEEbiography}[{\includegraphics[width=1in,height=1.35in,clip,keepaspectratio]{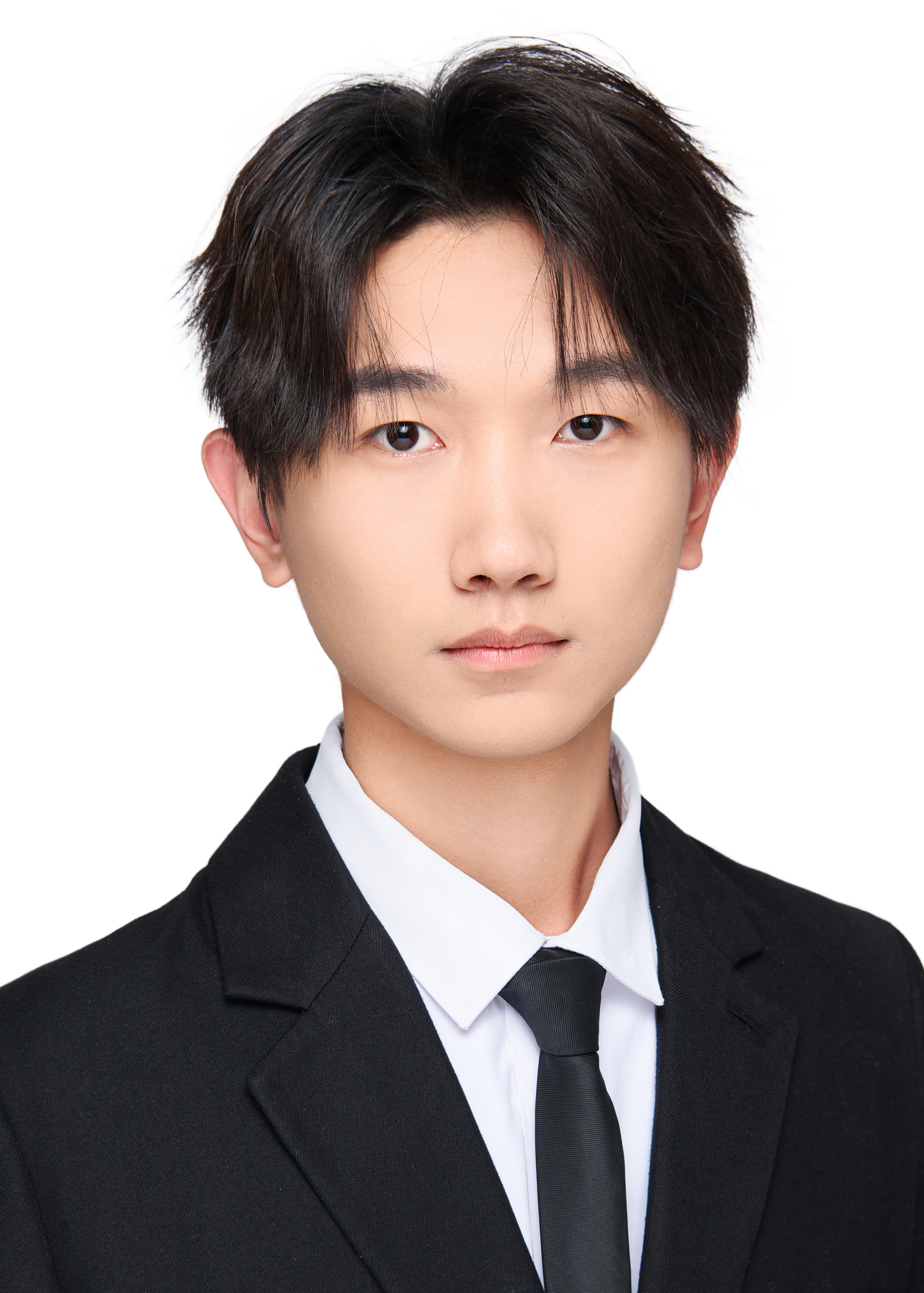}}]{Haoyuan Xuan} is current a undergraduate student studied in the School of Computer Science and Engineering at Southeast University, Nanjing, China. 
	His research interests include multi-label learning and weakly supervised learning.
	
	His research interests include multi-label learning and weakly  supervised learning.
\end{IEEEbiography}

\begin{IEEEbiography}[{\includegraphics[width=1in,height=1.25in,clip,keepaspectratio]{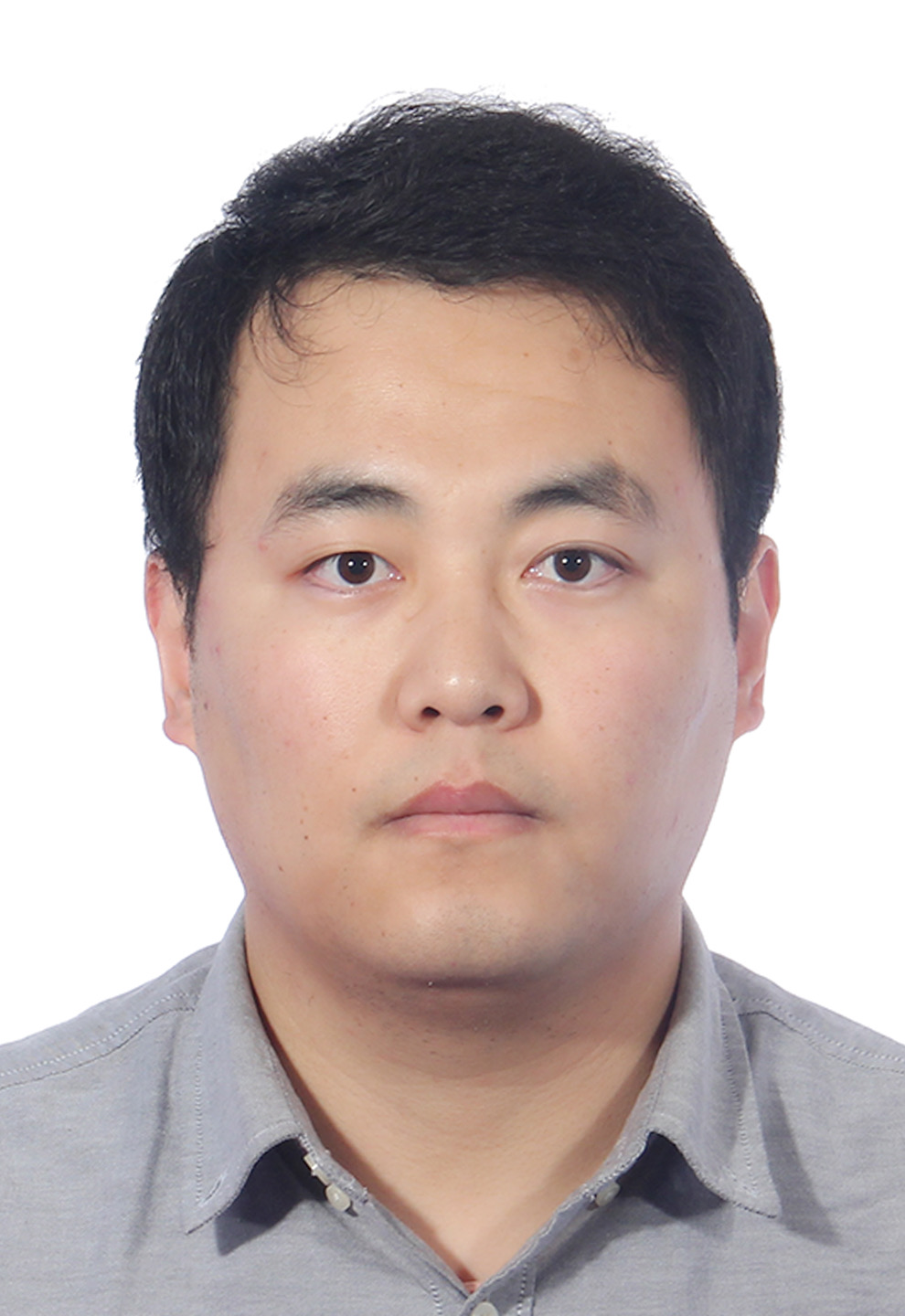}}]{Jing Wang} received the B.Sc. degree in computer science from Suzhou University of Science and Technology, Suzhou, China, in 2013, and the M.Sc. degree in computer science from Northeastern University, Shenyang, China, in 2015, and the Ph.D. degree in software engineering from Southeast University, Nanjing, China, in 2021. 
	
	He is currently an assistant professor  of the School of Computer Science and Engineering, Southeast University, Nanjing. His research interests include pattern recognition and machine learning. He has published over 10 refereed articles in these areas. 
\end{IEEEbiography}

\begin{IEEEbiography}[{\includegraphics[width=1in,height=1.25in,clip,keepaspectratio]{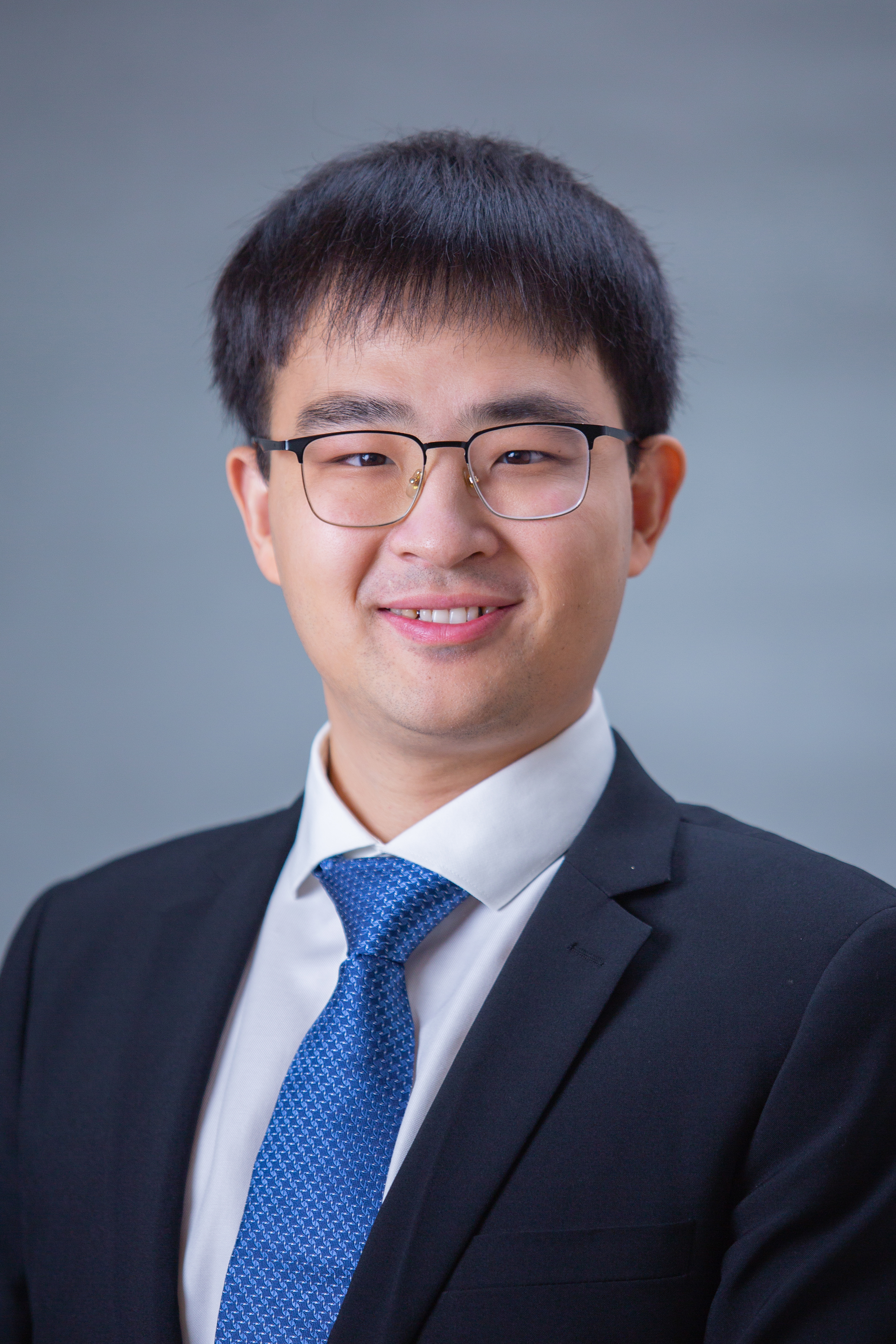}}]{Yuheng Jia}  (Member, IEEE) received the B.S. degree in automation and the M.S. degree in control theory and engineering from Zhengzhou University, Zhengzhou, China, in 2012 and 2015, respectively, and the Ph.D. degree in computer science from the City University of Hong Kong, Hong Kong, China, in 2019. He is currently an Associate Professor with the School of Computer Science and Engineering, Southeast University, Nanjing, China. His research interests broadly include topics in machine learning and data representation, such as weakly-supervised learning, high-dimensional data modeling and analysis, and low-rank tensor/matrix approximation and factorization.
 \end{IEEEbiography}

\begin{IEEEbiography}[{\includegraphics[width=1in,height=1.25in,clip,keepaspectratio]{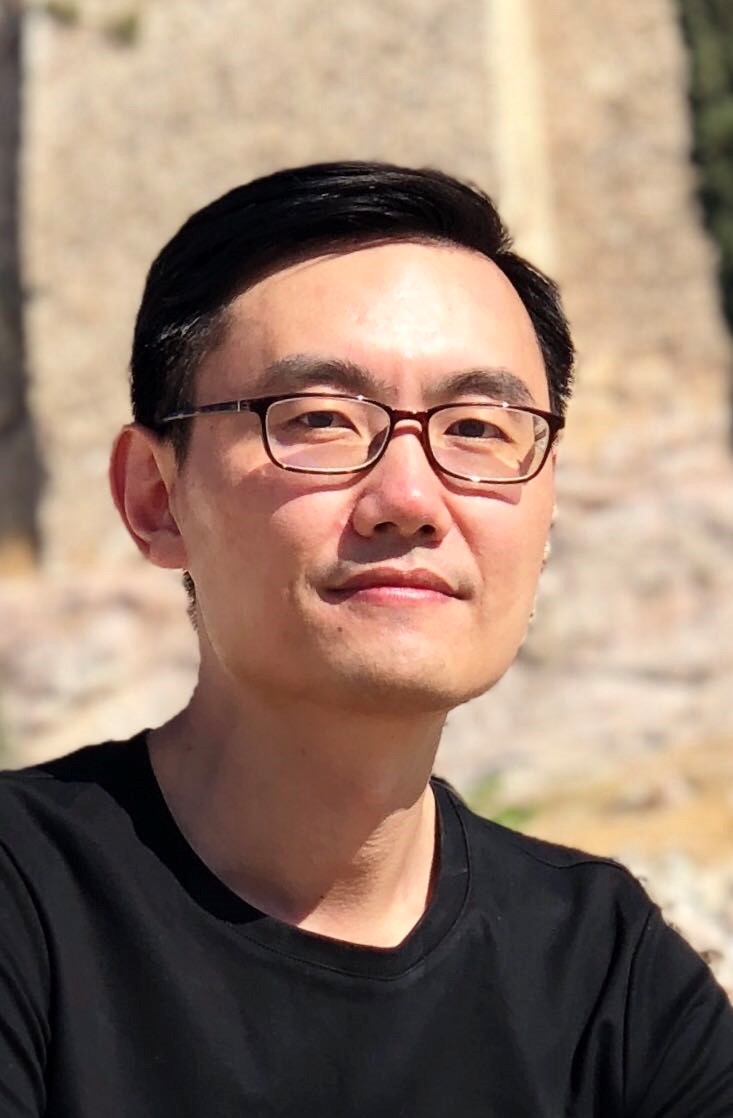}}]{Xin Geng}(Senior Member, IEEE) received the B.Sc. and M.Sc. degrees in computer science from Nanjing University, Nanjing, China, in 2001 and and 2004, respectively, and the Ph.D. degree in computer science from Deakin University, Geelong, VIC, Australia, in 2008. 
	
	He is currently a chair professor of the School of Computer Science and Engineering, Southeast University, Nanjing. His research interests include machine learning, pattern recognition, and computer vision. He has published over 100 refereed articles in these areas, including those published in prestigious journals and top international conferences. 
	
	Dr. Geng has been an Associate Editor of IEEE TRANSACTIONS ON MULTIMEDIA, Frontiers of Computer Science, and Mathematical Foundations of Computing, a Steering Committee Member of Pacific Rim International Conferences on Artificial Intelligence (PRICAI), a Program Committee Chair for conferences, such as PRICAI 2018 and  Vision And Learning SEminar (VALSE) 2013, the Area Chair for conferences, such as Computer Vision and Pattern Recognition (CVPR), ACM Multimedia, and Chinese Conference on Pattern Recognition (CCPR), and a Senior Program Committee Member for conferences, such as International Joint Conference on Artificial Intelligence (IJCAI), AAAI Conference on Artificial Intelligence (AAAI), and European Conference on Artificial Intelligence (ECAI). He is a Distinguished Fellow of International Engineering and Technology Institute. 
\end{IEEEbiography}

\end{document}